\documentclass{article} 
\usepackage{collas2025_conference,times}
\usepackage{easyReview}

\usepackage{amsmath,amsfonts,bm}

\def\1{\bm{1}}

\DeclareMathAlphabet{\mathsfit}{\encodingdefault}{\sfdefault}{m}{sl}
\SetMathAlphabet{\mathsfit}{bold}{\encodingdefault}{\sfdefault}{bx}{n}

\newcommand{\E}{\mathbb{E}}

\newcommand{\N}{\mathcal{N}_d}
\newcommand{\R}{\mathbb{R}}

\newcommand{\sigmoid}{\sigma}

\newcommand{\baralpha}{\bar{\alpha}}
\newcommand{\hatu}{\hat{u}}

\usepackage{natbib}
\usepackage{url}
\usepackage{graphicx}
\usepackage{subcaption}
\usepackage{booktabs}
\usepackage{multirow}

\usepackage{amsmath}
\usepackage{amssymb}
\usepackage{derivative}
\usepackage{float}

\usepackage{algorithm}
\usepackage{algpseudocode}
\usepackage[normalem]{ulem}
\usepackage{xcolor}

\usepackage{hyperref}
\hypersetup{
    colorlinks=true,
    linkcolor=red,
    filecolor=magenta,
    urlcolor=blue,
    citecolor=purple,
    pdftitle={Overleaf Example},
    pdfpagemode=FullScreen,
    }

\title{NoProp: Training Neural Networks without Full Back-propagation or Full Forward-propagation}

\author{Qinyu Li\\
Department of Statistics\\
University of Oxford\\
\texttt{qinyu.li@stats.ox.ac.uk} \\
\And
Yee Whye Teh \\
Department of Statistics\\
University of Oxford\\
\texttt{y.w.teh@stats.ox.ac.uk} \\
\And 
Razvan Pascanu  \\
Mila \\
\texttt{r.pascanu@gmail.com} \\
}

\collasfinalcopy 

\begin{document}

\maketitle

\begin{abstract}


    The canonical deep learning approach for learning requires computing a
    gradient term at each block by back-propagating the error signal from the
    output towards each learnable parameter.
    Given the stacked structure of neural networks, where each block builds on
    the representation of the block below, this approach leads to hierarchical
    representations.  More abstract features live on the top blocks of the
    model, while features on lower blocks are expected to be less abstract.
    In contrast to this, we introduce a new learning method named
    \emph{NoProp}, which does not rely on either forward or backwards
    propagation across the entire network.
    Instead, \emph{NoProp} takes inspiration from diffusion and flow matching
    methods, where each block independently learns to denoise a noisy target using only local targets and back-propagation within the block.
    We believe this work takes a first step towards introducing a new family of learning methods, that does not learn hierarchical
    representations -- at least not in the usual sense.
    \emph{NoProp} needs to fix the representation at each block beforehand to a noised version
    of the target, learning a local denoising process that can then be exploited at
    inference.
    We demonstrate the effectiveness of our method on MNIST, CIFAR-10, and
    CIFAR-100 image classification benchmarks.
    Our results show that \emph{NoProp} is a viable learning algorithm, is easy to use and computationally efficient.
    By departing from the traditional learning paradigm which requires back-propagating a global error signal, \emph{NoProp}
    alters how credit assignment is done within the network, enabling more efficient
    distributed learning as well as potentially impacting other characteristics of the learning process.
\end{abstract}

\section{Introduction}

Back-propagation~\citep{rumelhart1986learning}~has long been a cornerstone of
deep learning, and its application has enabled deep learning technologies to
achieve remarkable successes across a wide range of domains from sciences to
industries. Briefly, it is an iterative algorithm, which adapts the
parameters of a multi-block neural network at each step such that its output
match better a desired target. In this paper, we define a \emph{block} as either a single layer or a group of consecutive layers within a neural network. Each step of back-propagation first performs a
forward-propagation of the input signal to generate a prediction, then compares
the prediction to a desired target, and finally propagates the error signal back
through the network to determine how the weights of each block should be
adjusted to decrease the error.  This way, each block can be thought of as
learning to change the representation it receives from the block below it to
one that subsequent blocks can use to make better predictions. The error signal
propagated backwards is used to do \emph{credit assignment}, i.e. to decide how
much each parameter needs to change in order to minimize the error.

The simplicity of back-propagation has made it the de facto method for training
neural networks. However over the years there has been consistent interest in
developing alternative methods that do not rely on back-propagation. This
interest is driven by several factors. Firstly, back-propagation is
biologically implausible, as it requires synchronised alternation between
forward and backward passes \citep[e.g][]{lee2015difference}. Secondly,
back-propagation requires storing intermediate activations during the forward
pass to facilitate gradient computation in the backward pass, which can impose
significant memory overheads \citep{rumelhart1986learning}. Finally, the
sequential propagation of gradients introduces dependencies that impede
parallel computation, making it challenging to effectively utilize multiple
devices and servers for large-scale machine learning
\citep{carreira14distributed}. This sequential nature of how the credit
assignment is computed has also additional implications for learning, leading
to interference~\citep{schaul2019ray} and playing a role in catastrophic
forgetting~\citep{hadsell2020embracing}.

Alternative optimization methods to back-propagation include gradient-free
methods (e.g., direct search methods~\citep{fermi52, torczon91} and model-based methods~\citep{bortzKelley97, conn00}), zero-order gradient
methods~\citep{flaxman2004online,  duchi2015optimal, Nesterov2015,
Liu20,ren2022scaling},
evolution strategies \citep{wierstra2014natural,
salimans2017evolution,Such2017DeepNG,khadka2018evolution}, methods that rely
on local losses such as difference target propagation \citep{lee2015difference}
and the forward-forward algorithm \citep{hinton2022forward}. However, such approaches often retain standard neural network architectures and seek only to modify the optimization algorithm, and frequently fall short of back-propagation in terms of accuracy, computational efficiency and scalability.

In this paper, we explore an alternative path, rethinking both the architecture and the training process. We propose a novel training framework that avoids global back-propagation across network blocks, based on the denoising score matching approach that
underlies diffusion models
\citep{sohlDickstein15,ho2020denoising,song21score}. Unlike prior methods that approximate or replace back-propagation within standard architectures, NoProp employs a block-wise design where inputs are broadcast to each block, allowing each block to be trained independently using only local back-propagation. In brief, at
training time each block is trained to predict the target label given a noisy
label and the training input using back-propagation within the block, while at inference time each block takes the
noisy label produced by the previous block, and denoises it by taking a step
towards the label it predicts. This design avoids full forward and backward passes across the entire network, enabling block-local updates with minimal coordination, such as shared embeddings. While NoProp still uses back-propagation within blocks, its local training structure makes it more biologically plausible than full back-propagation and naturally suited for parallel and distributed training. We evaluate NoProp on MNIST, CIFAR-10, and CIFAR-100, demonstrating performance comparable to full back-propagation within the architecture we study. We also show that NoProp achieves better performance and efficiency than the adjoint sensitivity method used for training neural ordinary differential equations (Neural ODEs) \citep{Chen18}. While direct comparisons to other methods are limited by architectural differences, our aim is to show that local learning strategies inspired by generative models can be effectively applied to supervised learning. Although we have not fully explored scaling NoProp to larger datasets or more complex domains, we view NoProp as an early step toward more efficient, biologically-inspired, modular, and parallelizable alternatives to end-to-end back-propagation.

\section{Methodology}
In this section we will describe the NoProp method for learning without full forward- or back-propagation. While the technical ideas are an application of variational diffusion models \citep{sohlDickstein15,kingma2021variational,gulrajani2024likelihood}, we apply them in the context of alternative methods to end-to-end back-propagation, and interpret them differently as enabling learning without full forward or backward passes.
\subsection{NoProp}
Let $x$ and $y$ be a sample input-label pair in our classification dataset, assumed drawn from a data distribution $q_0(x,y)$, and $z_0,z_{1},\ldots,z_T \in \R^{d}$ be corresponding \emph{stochastic} intermediate activations of a neural network with $T$ blocks, which we aim to train to estimate $q_0(y|x)$.

\begin{figure}[tbp]
    \centering
    \includegraphics[width=\textwidth]{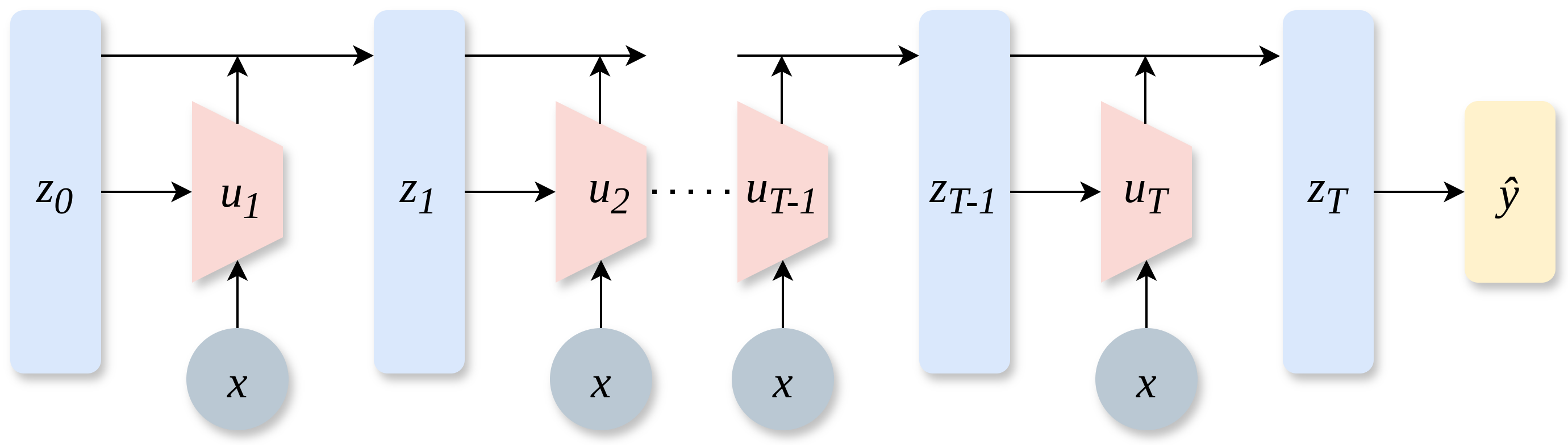} 
    \caption{Architecture of NoProp. $z_0$ represents Gaussian noise, while $z_1, \dots, z_T$ are successive transformations of $z_0$ through the learned dynamics $u_1, \dots, u_T$, with each block conditioned on the image $x$, ultimately producing the class prediction $\hat{y}$.}
    \label{fig:architecture}
\end{figure}

We define two distributions $p$ and $q$ decomposed in the following ways:
\begin{align}
    p((z_t)_{t=0}^T,y|x) &= \textstyle
    p(z_0)\left(\prod_{t=1}^{T} p(z_{t}|z_{t-1},x)\right) p(y|z_T), \\
    q((z_t)_{t=0}^T|y,x) &= \textstyle
    q(z_T|y)\left(\prod_{t=T}^1 q(z_{t-1}|z_{t})\right).
\end{align}
The distribution $p$ can be interpreted as a stochastic forward propagation process which iteratively computes the next activation $z_t$ given the previous one $z_{t-1}$ and the input $x$. In fact, we shall see that it can be explicitly given as a residual network with Gaussian noise added to the activations:
\begin{align}
    z_{t} &=
    a_t \hatu_{\theta_t}(z_{t-1},x) +
    b_t z_{t-1} +
    \sqrt{c_t} \epsilon_t, \quad
    \epsilon_t \sim \N(\epsilon_t | 0,1)
\label{eq:resblock}
\end{align}
where $\N(\cdot|0, 1)$ is a $d$-dimensional Gaussian density with mean vector 0 and identity covariance matrix,
$a_t, b_t, c_t$ are scalars (given below), $b_tz_{t-1}$ is a weighted skip connection, and $\hatu_{\theta_t}(z_{t-1},x)$ being a residual block parameterised by $\theta_t$.
Note that this computation structure is different from a standard deep neural network which does not have direct connections from the input $x$ into each block.

Following the variational diffusion model approach, we can alternatively interpret $p$ as a conditional latent variable model for $y$ given $x$, with the $z_t$'s being a sequence of latent variables. We can learn the forward process $p$ using a variational formulation, with the $q$ distribution serving as the variational posterior. The objective of interest is then the ELBO, a lower bound of the log likelihood $\log p(y|x)$ (a.k.a.\ the evidence):
\begin{align}
\log p(y|x) &\ge
\E_{q((z_t)_{t=0}^T|y,x)} \left[ \log p((z_t)_{t=0}^T,y|x) - \log q((z_t)_{t=0}^T|y,x) \right].
\label{eq:elbo}
\end{align}
See Appendix~\ref{sec:elbo} for the derivation of Equation~\ref{eq:elbo}. Following \citet{sohlDickstein15,kingma2021variational}, we fix the variational posterior $q$ to a tractable Gaussian distribution; here we use the variance preserving Ornstein-Uhlenbeck process:
\begin{align}
q(z_T|y) &= \N(z_T|\sqrt{\alpha_T}u_y,1-\alpha_T), &
q(z_{t-1}|z_{t}) &= \N(z_{t-1}|\sqrt{\alpha_{t-1}}z_{t},1-\alpha_{t-1})
\end{align}
where $u_y$ is an embedding of the class label $y$ in $\R^d$, defined by a trainable embedding matrix $W_{\mathrm{Embed}} \in \mathbb{R}^{m \times d}$, with $m$ as the number of classes. The embedding is given by $u_y = \{W_{\text{Embed}}\}_{y}$. Using standard properties of the Gaussian, we can obtain
\begin{align}
q(z_t|y) &= \N(z_t|\sqrt{\baralpha_t}u_y,1-\baralpha_t), &
q(z_{t}|z_{t-1},y) &= \N(z_{t}|\mu_t(z_{t-1},u_y),c_t)
\end{align}
with $\baralpha_t = \prod_{s=t}^T \alpha_s $,
$\mu_t(z_{t-1},u_y) =
a_t u_y +
b_t z_{t-1}$,
$a_t=\frac{\sqrt{\baralpha_{t}}(1-\alpha_{t-1})}{1-\baralpha_{t-1}}$,
$b_t=\frac{\sqrt{\alpha_{t-1}}(1-\baralpha_{t})}{1-\baralpha_{t-1}}$, and
$c_t =
\frac{(1-\baralpha_{t})(1-\alpha_{t-1})}{1-\baralpha_{t-1}}
$. See Appendix~\ref{sec:diffusion_noising} and \ref{sec:diffusion_denoising} for a full derivation. To optimise the ELBO, we parameterise $p$ to match the form for $q$:
\begin{align}
    p(z_{t}|z_{t-1},x) &= \N(z_{t}|\mu_t(z_{t-1},\hatu_{\theta_t}(z_{t-1},x)), c_t), &
    p(z_0) &= \N(z_0|0,1)
\end{align}
where $p(z_0)$ has been chosen to be the stationary distribution of the Ornstein-Uhlenbeck process, and $\hatu_{\theta_t}(z_{t-1},x)$ is a neural network block parameterised by $\theta_t$. The resulting computation to sample $z_{t}$ given $z_{t-1}$ and $x$ is as given in by the residual architecture (Equation \ref{eq:resblock}), with
$a_t, b_t, c_t$ as specified above. Finally, plugging this parameterisation into the ELBO (Equation \ref{eq:elbo}) and simplifying, we obtain the NoProp objective,
\begin{align}
    \mathcal{L}_{\textrm{NoProp}} = &\E_{q(z_{T}|y)}\left[ -\log \hat{p}_{\theta_{\textrm{out}}}(y|z_T) \right] \nonumber \\
    &+ D_{\mathrm{KL}}(q(z_0 | y) \Vert p(z_0)) \nonumber \\
    &+ \frac{T}{2} \eta \E_{t \sim \mathcal{U}\{1, T\}} \left[(\textrm{SNR}(t) - \textrm{SNR}(t-1)) \| \hatu_{\theta_t}(z_{t-1},x) - u_y \| ^2\right], \label{eq:noprop_dt}
\end{align}
where $\textrm{SNR}(t) = \frac{\baralpha_{t}}{1 - \baralpha_{t}}$ is the \emph{signal-to-noise ratio}, $\eta$ is a hyperparameter, and $\mathcal{U}\{1, T\}$ is the uniform distribution on the integers $1, \ldots, T$. See Appendix~\ref{sec:objectives} for a full derivation of the NoProp objective.

We see that each $\hatu_{\theta_t}(z_{t-1},x)$ is trained to directly predict $u_y$ given $z_{t-1}$ and $x$ with an L2 loss, while $\hat{p}_{\theta_{\textrm{out}}}(y|z_T)$ is trained to minimise cross-entropy loss. Each block $\hatu_{\theta_t}(z_{t-1},x)$ is trained independently, and this is achieved without full forward- or back-propagation across the entire network.

\subsection{Variations}

\subsubsection{Fixed and learnable $W_{\mathrm{Embed}}$}
The class embedding matrix $W_{\mathrm{Embed}}$ can be either fixed or jointly learned with the model. In the fixed case, we set $W_{\mathrm{Embed}}$ to be the identity matrix, where each class embedding is a one-hot vector. In the learned case, $W_{\mathrm{Embed}}$ is initialized as an orthogonal matrix when possible, or randomly otherwise. When the embedding dimension matches the image dimension, we interpret this as learning an ``image prototype." In this special case, class embeddings are initialized as the images with the smallest median distance to all other images within the same class, serving as crude prototypes before training.

\subsubsection{Continuous-time diffusion models and neural ODEs}
The NoProp framework presented above involves learning $\hatu_{\theta_t}(z_{t-1},x)$ parameterised by $\theta_t$ for each time step $t = 1, 2, \ldots, T$ along with a linear layer for $\hat{p}_{\theta_{\textrm{out}}}(y|z_T)$. We can extend this to a continuous version where the number of latent variables $T$ tends to infinity \citep{kingma2021variational}, leading to the objective
\begin{equation} \label{eq:noprop_ct}
    \E_{q(z_1|y)}\left[ -\log \hat{p}_{\theta_{\textrm{out}}}(y|z_1) \right]
    + D_{\mathrm{KL}}(q(z_0 | y) \Vert p(z_0))
    + \frac{1}{2} \eta \E_{t \sim \mathcal{U}[0, 1]} \left[\textrm{SNR}'(t) \| \hatu_{\theta}(z_{t},x, t) - u_y \| ^2\right].
\end{equation}
See Appendix~\ref{sec:objectives} for a full derivation. Here, time $t$ is treated as a continuous variable in $[0, 1]$. We use a continuous noise schedule and $\textrm{SNR}'(t)$ denotes the derivative of the signal-to-noise ratio. A single block $\hatu_{\theta}(z_{t},x, t)$ is trained for all time steps $t$, with $t$ as an additional input. The evolution of latent variables \( z_t \) follows a continuous-time diffusion process, described by a stochastic differential equation (SDE). Note that there exits a corresponding ordinary differential equation (ODE)
$    \frac{d}{dt} z_t = f(z_t | x, t)
$
sharing the same marginal distributions as the SDE \citep{song21score}. The function \( f(z_t | x, t) \) represents the deterministic vector field generating \( z_t \)'s evolution. Neural networks parameterised in this manner are known as neural ODEs \citep{Chen18}.

Neural ODE training usually relies on back-propagating through the ODE solver to optimize task-specific loss functions (e.g., cross-entropy in classification), or using the adjoint sensitivity method \citep{Chen18} which estimates the gradient by solving another ODE backward in time. The continuous-time diffusion model instead learns an ODE dynamic that inverts a predefined noising process. Training occurs by sampling time steps independently, without requiring full forward or backward passes through time. This makes training more efficient while still enabling expressive ODE dynamics. 

\subsubsection{Flow matching}
An alternative approach to NoProp’s continuous formulation is flow matching \citep{lipman2022flow,tong2023improving}, which directly learns the vector field $f(z_t | x, t)$ that transports noise toward the predicted label embedding via an ODE. Since $f(z_t | x, t)$ is generally unknown, flow matching instead learns the conditional vector field $f(z_t | z_0, z_1, x, t)$, where $z_0$ is the initial noise and $z_1 = u_y$ is the label embedding. The conditional vector field is user-specified. A simple choice is to define a Gaussian probability path between noise $z_0$ and label embedding $z_1 = u_y$
\begin{equation}
p_t(z_t | z_0, z_1, x) = \N(z_t \mid tz_1 + (1-t) z_0, \sigma^2),
\end{equation}
with the corresponding vector field $f(z_t | z_0, z_1, x, t) = z_1 - z_0$. The flow matching objective is then
\begin{equation} \label{eq:noprop_fm}
    \mathbb{E}_{t \sim \mathcal{U}[0, 1], q_0(x, y), p(z_0), p_t(z_t |z_0, z_1, x)} \lVert v_{\theta}(z_t, x, t) - (z_1 - z_0) \rVert^2,
\end{equation}
where $v_{\theta}(z_t, x, t)$ is a neural network with parameters $\theta$. When the label embeddings are jointly learned, we introduce an additional anchor loss to prevent different class embeddings from collapsing. Following prior work \citep{gao2022difformer, hu2024flow}, we incorporate a cross-entropy term \(- \log \hat{p}_{\theta_{\mathrm{out}}}(y \mid \tilde{z}_1(z_t,x,t))\), where \(\tilde{z}_1(z_t,x,t)\) is an extrapolated linear estimate
\begin{equation}
    \tilde{z}_1(z_t,x,t) = z_t + (1-t) v_{\theta}(z_t, x, t).
\end{equation}
The modified flow matching objective is
\begin{equation}
\label{eq:noprop_fm_anchor}
    \mathbb{E}_{t \sim \mathcal{U}[0, 1], q_0(x, y), p(z_0), p_t(z_t |z_0, z_1, x)}  [\lVert v_{\theta}(z_t, x, t) - (z_1 - z_0) \rVert^2
     - \log \hat{p}_{\theta_{\mathrm{out}}}(y \mid \tilde{z}_1(z_t,x,t))].
\end{equation}
Similarly to diffusion, we parameterise $\hat{p}_{\theta_{\mathrm{out}}}(y \mid \tilde{z}_1(z_t,x,t))$ using a linear layer followed by softmax. This formulation ensures well-separated class embeddings. 

\subsection{Implementation details}
\subsubsection{Architecture}
The NoProp architecture is illustrated in Figure~\ref{fig:architecture}. During inference, Gaussian noise \( z_0 \) undergoes a sequence of transformations through the diffusion process. At each step, the latent variable \( z_t \) evolves via a diffusion dynamic block \( u_t \), producing a sequence \( z_1, z_2, \dots, z_T \) until reaching \( z_T \). Each $u_t$ is conditioned on the previous latent state $z_{t-1}$ and the input image \( x \). Finally, a linear layer followed by a softmax function maps \( z_T \) to the predicted label \( \hat{y} \). The model used to parameterise each $u_t$ block is described in Section~\ref{sec:model}.

This architecture is designed specifically for inference. During training, time steps are sampled, and each diffusion dynamic block \( u_t \) is trained independently. The linear layer and the embedding matrix are trained jointly with the diffusion blocks, as the linear layer helps prevent the class embeddings from collapsing.

For the flow matching variant, the $u_t$'s in Figure~\ref{fig:architecture} represent the ODE dynamics. The label prediction \( \hat{y} \) is obtained directly from \( z_T \) by finding the class embedding closest to \( z_T \) in terms of Euclidean distance.

\subsubsection{Training procedure} \label{sec:model}

\paragraph{Models} 
The models used for training are illustrated in Figure~\ref{fig:models} in the Appendix. For discrete-time diffusion, we use a neural network $\hat{u}_{\theta_t}$ to model the diffusion dynamics \( u_{t} \). The model takes an input image \( x \) and a latent variable \( z_{t-1} \) in the label embedding space, processing them through separate embedding pathways before concatenation. The image \( x \) is passed through a convolutional embedding module followed by a fully connected layer. When the embedding dimension matches the image dimension, \( z_{t-1} \) is treated as an image and embedded similarly using a convolutional module. Otherwise, \( z_{t-1} \) is passed through a fully connected network with skip connections. The fused representation is then processed by additional fully connected layers, producing logits. We then apply a softmax function to the logits, yielding a probability distribution over class embeddings. The final output of $\hat{u}_{\theta_t}$ is then computed as a weighted sum of the class embeddings in \( W_{\mathrm{Embed}} \) using this probability distribution.

For continuous-time diffusion, we train \( \hat{u}_{\theta} \), which takes an additional timestamp \( t \) as input. The timestamp is encoded using positional embeddings and processed through a fully connected layer before being concatenated with those of \( x \) and \( z_t \). The rest of the model follows roughly the same structure as in the discrete case. 

For flow matching, we train a neural network \( \hat{v}_{\theta} \). The architecture remains the same as in the continuous-time diffusion case, but instead of applying softmax, \( \hat{v}_{\theta} \) is obtained by directly computing the weighted sum of the class embeddings using the logits, treating the logits as unrestricted weights. This allows \( \hat{v}_{\theta} \) to represent arbitrary directions in the embedding space, unlike \( \hat{u}_{\theta} \), which is constrained to a convex combination of class embeddings.

While the overall model structure remains similar across settings, the inclusion of \( t \) in the continuous-time diffusion and flow matching cases introduces an additional input and processing step compared to the discrete-time diffusion case. We choose to parameterise $\hat{p}_{\theta_{\textrm{out}}}(y|z_T)$ in Equation~\ref{eq:noprop_dt}, $\hat{p}_{\theta_{\textrm{out}}}(y|z_1)$ in Equation~\ref{eq:noprop_ct}, and $\hat{p}_{\theta_{\mathrm{out}}}(y \mid \tilde{z}_1(z_t, x, t))$ in Equation~\ref{eq:noprop_fm_anchor} using a linear layer followed by softmax. 

\paragraph{Noise schedule}
For discrete-time diffusion, we use a fixed cosine noise schedule. For continuous-time diffusion, we train the noise schedule jointly with the model. Further details on the trainable noise schedule are provided in Appendix~\ref{sec:noise_schedule}.

\section{Related Works}
\subsection{Alternative Methods to Back-propagation}

\begin{figure}[tbp]
    \centering
    \includegraphics[width=\textwidth]{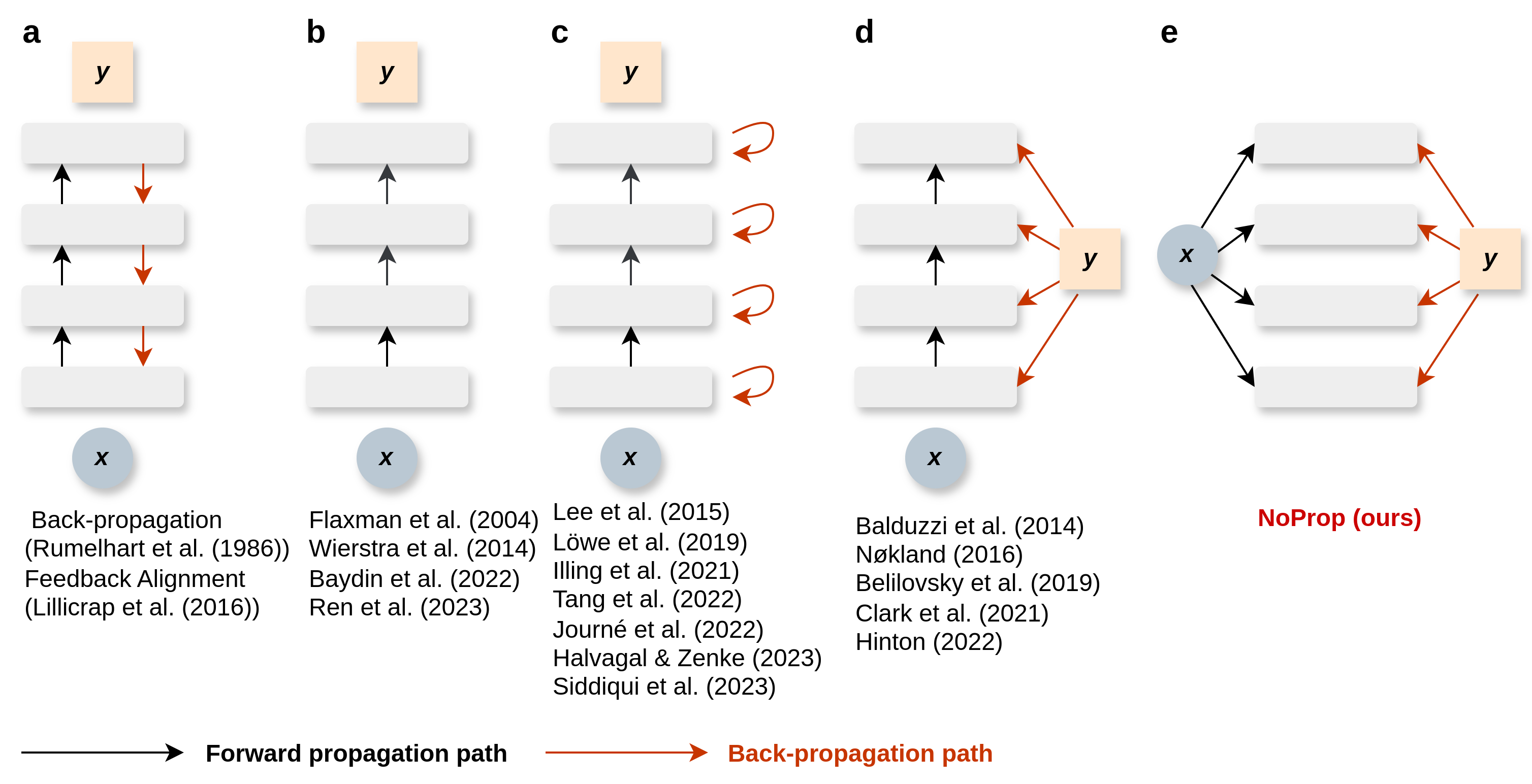} 
    \caption{Comparison of weight update strategies in end-to-end back-propagation and its alternatives. \textbf{a} \textit{Feedback Alignment} closely resembles standard end-to-end back-propagation. \textbf{b} A variety of optimization methods that eliminate the need for backward passes entirely. \textbf{c} After a forward pass, each block updates its weights using local unsupervised or self-supervised objectives, while previous blocks are kept frozen. \textbf{d} Similar to \textbf{c}, but the target labels are broadcast to each block, enabling local supervised updates. \textbf{e} \textit{NoProp} broadcasts both the input and the labels to every block, allowing each to update independently using only local back-propagation.}
    \label{fig:backprop_alternatives}
\end{figure}

A wide range of alternatives to back-propagation have been proposed, aiming either to improve biological plausibility, increase computational efficiency, or both. Below, we highlight key approahces and illustrate their weight updates in Figure~\ref{fig:backprop_alternatives}. One well-known limitation of backpropagation is the ``weight transport problem'' \citep{grossberg1987competitive}, which refers to the biologically implausible requirement that the backward pass must have access to the exact weights used in the forward pass. Feedback Alignment \citep{Lillicrap2016RandomSF} addresses this by replacing the backward weights with fixed, random matrices. Although the network still relies on a top-down propagation of errors, this method avoids symmetric weight transport, making it more biologically plausible. However, it does not reduce computational cost since it still requires sequential backward passes.

Several optimization approaches avoid backward passes entirely by using only forward computations to update parameters. These include zeroth-order gradient estimators \citep{flaxman2004online, duchi2015optimal, Nesterov2015, Liu20} and evolutionary strategies \citep{wierstra2014natural, salimans2017evolution, khadka2018evolution}. Evolution strategies are inspired by natural selection and operate without gradient information, making them well-suited for black-box optimization. However, they are typically sample-inefficient, especially in high-dimensional spaces, and require large numbers of function evaluations. A related but more efficient alternative is forward gradient descent \citep{baydin2022gradientsbackpropagation}, which uses only forward passes to compute an unbiased estimate of the gradient and reduces computational cost compared to back-propagation.

Another prominent class of methods focuses on block-wise local learning, where the network is divided into modules or blocks that are trained independently using local objectives. These methods eliminate the need for error signals to propagate through the entire network and are naturally amenable to parallel training. Early examples of this idea appear in deep belief networks \citep{hinton2006fast}. In more recent work, Direct Feedback Alignment \citep{nokland2016directfeedbackalignmentprovides} extends the original Feedback Alignment idea by applying independent random feedback paths to each block. Greedy layerwise training with auxiliary networks \citep{belilovsky2019greedylayerwiselearningscale} appends auxiliary networks to each block to optimize a supervised learning objective for that block. \citet{clark2021creditassignmentbroadcastingglobal} propose broadcasting a global error signal to all hidden units, eliminating the need for unit-specific feedback. The Forward-Forward algorithm \citep{hinton2022forward} trains each block by contrasting its response to positive and negative examples.

An alternative branch of block-wise methods relies on unsupervised or self-supervised local objectives. These approaches update each block independently to learn useful representations, which are later used for downstream supervised learning. Examples include Hebbian learning and local learning rules based on reconstruction, contrastive objectives, or predictive coding \citep{lowe2019putting, illing2021local, Tang_2022, journe2022hebbian, halvagal2023combination, siddiqui2023blockwise}. These methods maintain biological plausibility through local updates, and they are computationally efficient due to their potential for parallelism.

Several other approaches also decouple updates between layers or blocks. \citet{ren2022scaling} extend forward gradient descent to a block-wise local training setting. Synthetic gradients \citep{jaderberg2017decoupledneuralinterfacesusing} train small networks to predict the gradient at each layer, allowing for parallel updates. Kickback \citep{balduzzi2014kickbackcutsbackpropsredtape} simplifies the gradient computation by retaining only a global error signal. Target propagation \citep{lee2015difference}replaces error back-propagation with a learned inverse function that propagates target activations back through the network. While these methods can increase biological plausibility, their effectiveness often depends on the quality of the learned components, such as the synthetic gradient models or inverse mappings.

Our method, NoProp, builds on the idea of block-wise local learning but differs in key ways. Both the input (e.g., an image in classification tasks) and the label are broadcast to each block of the network. The forward pass is predetermined by a noise schedule, so a full forward pass through the entire network is not required. Each block is trained with a local objective using back-propagation internally, without relying on end-to-end gradient flow. When label embeddings are fixed (e.g., using one-hot vectors), blocks can be updated independently, enabling parallel training and improving computational efficiency relative to standard back-propagation. If the label embeddings or noise schedule are jointly learned with the model, some limited coupling between blocks is introduced, but the method still avoids global back-propagation. In continuous-time settings, NoProp is also more efficient than approaches requiring ODE simulations, such as the adjoint sensitivity method \citep{Chen18}. Finally, because NoProp does not rely on propagating a global error signal across the network, it offers greater biological plausibility than end-to-end back-propagation.

\subsection{Diffusion and flow matching}
Several works in diffusion and flow matching are closely related to our method. \citet{han2022card} introduced classification and regression diffusion models, \citet{kim2025simulation} proposed flow matching for paired data, and \citet{hu2024flow} applied flow matching to conditional text generation. In contrast, our paper explores the implications of these ideas within the framework of alternative learning paradigms, building on the observation that diffusion and flow matching methods do not require full forward and backward passes.

\subsection{Representation Learning}

Generally speaking, most alternative methods to back-propagation, whether they
simply approximate gradients differently, or utilize a different search scheme
--- as it is the case of evolution strategies --- still rely on learning
intermediate representations that build on top of each other \citep{Bengio13Representations}. This allows learning more abstract representations as we look at
deeper blocks of the model, and is thought  to be of fundamental importance for deep learning and for representing cognitive processes~\citep{Markman2000}.  Indeed, initial successes of deep
learning have been attributed to the ability to train \emph{deep} architectures
to learn hierarchical representations~\citep{hinton2006fast, Bengio13Representations}, while early
interpretability work focused on visualising these increasingly more complex
features~\citep{Zeiler14, Lee09}.

By getting each block to learn to denoise labels, with the label noise
distribution chosen by the user, we can argue that NoProp does not \emph{learn}
global representations across blocks. Rather, it relies on representations \emph{designed} by the
user (specifically, the representations in intermediate blocks are Gaussian noised embeddings of target labels for diffusion, and an interpolation between Gaussian noise and embeddings of target labels for flow matching).
This is perhaps unsurprising: forward- and back-propagation can be
understood as information being disseminated across the blocks of a neural
network in order to enable the representation of each block to ``fit in'' with
those of neighbouring blocks and such that the target label can be easily
predicted from the representation at the last block. So for NoProp to work
without forward- or back-propagation, these intermediate representations have to be
fixed beforehand, i.e.\ to be designed by the user. 

The fact that NoProp achieves good performance without global representation learning
leads to the question of whether such learning is in fact necessary
for deep learning, and whether by designing representations we can enable
alternative approaches to deep learning with different characteristics.
Specifically, note that representations fixed in NoProp are not those one might think of as being more abstract in later blocks, which opens the door
to revisiting the role of hierarchical representations in modelling complex behaviour~\citep{Markman2000}.
These questions can become increasingly important as core assumptions of back-propagation based learning, like the i.i.d.\ assumption and sequential propagation of information
and error signal through the network, are proving to be limiting.

\section{Experiments}
We compare NoProp against back-propagation in the discrete-time case and against the adjoint sensitivity method \citep{Chen18} in the continuous-time case for image classification tasks. Details on the hyperparameters are provided in Table~\ref{table:experiment_details} in the Appendix.

\paragraph{NoProp (discrete-time)}

We refer to this method as NoProp with Discrete-Time Diffusion (NoProp-DT). We fix \( T = 10 \) and, during each epoch, update the parameters for each time step sequentially, as outlined in Algorithm~\ref{alg:noprop_dt}. While sequential updates are not strictly necessary for the algorithm to work — as time steps can also be sampled — we choose this approach to align with prior alternative methods to full back-propagation, ensuring consistency with existing approaches in the literature. In terms of parameterisation of the class probabilities $\hat{p}_{\theta_{\textrm{out}}}(y|z_T)$ in Equation~\ref{eq:noprop_dt}, we explore two approaches. The first, as described earlier, involves using softmax on the outputs of a fully connected layer $f$ with parameters $\theta_{\textrm{out}}$, such that
\begin{equation}
\hat{p}_{\theta_{\textrm{out}}}(y \mid z_T) = \text{softmax}(f_{\theta_{\mathrm{out}}}(z_T)) = \frac{\exp(f_{\theta_{\mathrm{out}}}(z_T)_y)}{\sum_{y'=1}^{m} \exp(f_{\theta_{\mathrm{out}}}(z_T)_{y'})},
\end{equation}
where $m$ is the number of classes. As an additional exploration, we investigate an alternative approach that relies on radial distance to parameterise the class probabilities. In this approach, we estimate a reconstructed label $\tilde{y}$ from $z_T$ by applying softmax to the output of the same fully connected layer $f_{\theta_{\mathrm{out}}}$, followed by a projection onto the class embedding matrix \( W_{\mathrm{Embed}} \), such that $\tilde{y}$ is a weighted embedding. The resulting probability of class \( y \) given $z_T$ is then based on the squared Euclidean distance between \( \tilde{y} \) and the true class embedding \( u_y = (W_{\mathrm{Embed}})_y\):
\begin{align}
\tilde{y} &= \text{softmax}(f_{\theta_{\mathrm{out}}}({z_T})) W_{\mathrm{Embed}}, \\
\hat{p}_{\theta_{\textrm{out}}}(y|z_T) &= \frac{\exp \left( -\frac{\| \tilde{y} - u_y \|^2}{2\sigma^2} \right)}{\sum_{y'=1}^{m} \exp \left( -\frac{\| \tilde{y} - u_{y'} \|^2}{2\sigma^2} \right)}.
\end{align}
This formulation can be interpreted as computing the posterior class probability assuming equal prior class probabilities and a normal likelihood $p(\tilde{y}|y) = \N(\tilde{y} | u_y, \sigma^2)$.

\paragraph{Back-propagation.} We also fix $T = 10$ and the forward pass is given by
\begin{align}
    z_0 & \sim \N(z_0 | 0, 1), \\
    z_t & = (1 - \alpha_t) z_{t-1} + \alpha_t \hat{u}_{\theta_t}(z_{t-1}, x), \quad t = 1, \dots, T,\\
    \hat{y} & = \operatorname{argmax}_{i \in \{1, \dots, m\}} \hat{p}_{\theta_{\mathrm{out}}}(y_i | z_T),
\end{align}
where \( \alpha_1, \dots, \alpha_T \) are learnable parameters constrained to the range \( (-1,1) \), defined as \( \alpha_t = \tanh(w_t) \) with learnable \( w_1, \dots, w_T \). This design closely resembles the forward pass of NoProp-DT, but with the network trained by standard back-propagation. The $\hat{u}_{\theta_t}(z_{t-1}, x)$ and $\hat{p}_{\theta_{\mathrm{out}}}(y | z_T)$ have identical model structures to those in NoProp-DT. 

\paragraph{NoProp (continuous-time)}
We refer to the continuous-time variants as NoProp with Continuous-Time Diffusion (NoProp-CT) and NoProp with Flow Matching (NoProp-FM). During training, the time variable \( t \) is randomly sampled from \( [0,1] \). During inference, we set \( T=1000 \) steps to simulate the diffusion process for NoProp-CT and the probability flow ODE for NoProp-FM. Detailed training procedures are provided in Algorithms~\ref{alg:noprop_ct} and \ref{alg:noprop_fm}.

\paragraph{Adjoint sensitivity}
We also evaluate the adjoint sensitivity method \citep{Chen18}, which trains a neural ODE using gradients estimated by solving a related adjoint equation backward in time. Including this method provides a meaningful baseline to assess the efficiency and accuracy of our approach in the continuous-time case. For fair comparison, we fix \( T=1000 \) during both training and inference and use a linear layer for final classification.

\begin{table}[]
    \centering
    \begin{tabular}{ccccccc}
    \toprule
    Method                 & \multicolumn{2}{c}{MNIST} & \multicolumn{2}{c}{CIFAR-10} & \multicolumn{2}{c}{CIFAR-100} \\ \midrule
                           & Train       & Test        & Train         & Test         & Train         & Test          \\ \midrule
    \multicolumn{7}{c}{Discrete-time}                                                                                 \\ \midrule
    Backprop (one-hot)     & 100.0±0.0   & 99.46±0.06  & 99.98±0.01    & 79.92±0.14   & 98.63±1.34    & 45.85±2.07    \\
    Backprop (dim=20)      & 99.99±0.0   & 99.43±0.03  & 99.96±0.02    & 79.3±0.52    & 94.28±7.43    & 46.57±0.87    \\
    Backprop (prototype)  & 99.99±0.01 & 99.44±0.05          & 99.97±0.01 & 79.58±0.44         & 99.19±0.71  & \textbf{47.8±0.19} \\
    NoProp-DT (one-hot)    & 99.92±0.01  & 99.47±0.05  & 95.02±0.19    & 79.25±0.28   & 84.97±0.67    & 45.93±0.46    \\
    NoProp-DT (dim=20)     & 99.93±0.01  & 99.49±0.04  & 94.95±0.09    & 79.12±0.37   & 83.25±0.39    & 45.19±0.22    \\
    NoProp-DT (prototype) & 99.97±0.0  & \textbf{99.54±0.04} & 97.23±0.11 & \textbf{80.54±0.2} & 90.7±0.14   & 46.06±0.25         \\ \midrule
    \multicolumn{7}{c}{Continuous-time}                                                                               \\ \midrule
    Adjoint (one-hot)      & 98.7±0.13   & 98.62±0.14  & 70.64±0.49    & 66.78±0.76   & 26.72±0.81    & 25.03±0.7     \\
    NoProp-CT (one-hot)    & 97.88±0.61  & 97.84±0.71  & 97.31±0.84    & 73.35±0.55   & 75.1±3.43     & 33.66±0.5     \\
    NoProp-CT (dim=20)     & 97.7±0.42   & 97.7±0.51   & 94.88±3.08    & 71.77±2.47   & 74.22±2.33    & 33.99±1.08    \\
    NoProp-CT (prototype) & 97.18±1.02 & 97.17±0.94          & 86.2±7.34  & 66.54±3.63         & 40.88±10.72 & 21.31±4.17         \\
    NoProp-FM (one-hot)    & 99.97±0.0   & 99.21±0.09  & 98.46±0.4     & 73.14±0.9    & 12.69±10.4    & 6.38±4.9      \\
    NoProp-FM (dim=20)     & 99.99±0.0   & \textbf{99.29±0.05}  & 99.49±0.15    & 73.5±0.28    & 83.49±4.62    & 31.14±0.52    \\
    NoProp-FM (prototype)  & 99.27±0.09  & 98.52±0.16  & 99.8±0.03     & \textbf{75.18±0.57}   & 96.37±1.09    & \textbf{37.57±0.32}    \\ \midrule
    \multicolumn{7}{c}{Previous methods without full back-propagation}                                                                \\ \midrule
    Forward-Forward        & -           & 98.63       & -             & -            & -             & -             \\
    Forward Gradient       & 100.00        & 97.45       & 80.61         & 69.32        & -             & -             \\
    Difference Target Prop & -           & 98.06       & -             & 50.71        & -             & -
    \end{tabular}
    \caption{Classification accuracies (\%) on MNIST, CIFAR-10, and CIFAR-100. Means and standard errors are computed from 15 values per method, obtained from 3 seeds and 5 inference runs per seed. Forward Gradient refers to the Local Greedy Forward Gradient Activity-Perturbed (LG-FG-A) method in \citet{ren2022scaling}. one-hot: fixed one-hot label embedding; dim=20: learned label embedding of dimension 20; prototype: learned label embedding of dimension equal to image size.}
    \label{table:results}
\end{table}

\begin{table}[]
    \centering
    \begin{tabular}{@{}lll@{}}
    \toprule
    \multicolumn{1}{c}{Method} & \multicolumn{2}{c}{Tiny ImageNet}                                     \\ \midrule
                               & \multicolumn{1}{c}{Train}          & \multicolumn{1}{c}{Test}         \\ \midrule
    Backprop (one-hot)         & 75.66±33.67 &  23.83±6.7 \\
    Backprop (prototype) &  99.82±0.02 &  \textbf{31.26±0.1}\\
    NoProp-DT (one-hot)        &  69.78±1.71  &  26.55±0.4 \\
    NoProp-DT (prototype)      &  78.45±0.97  & 25.65±0.19                       \\ \bottomrule
    \end{tabular}
    \caption{Classification accuracies (\%) on Tiny ImageNet in the discrete-time setting. Means and standard errors are computed from 15 values per method, obtained from 3 seeds and 5 inference runs per seed.}    
    \label{table:results_tiny_imagenet}
    \end{table}

\begin{figure}[tbp]
    \centering
    \includegraphics[width=\textwidth]{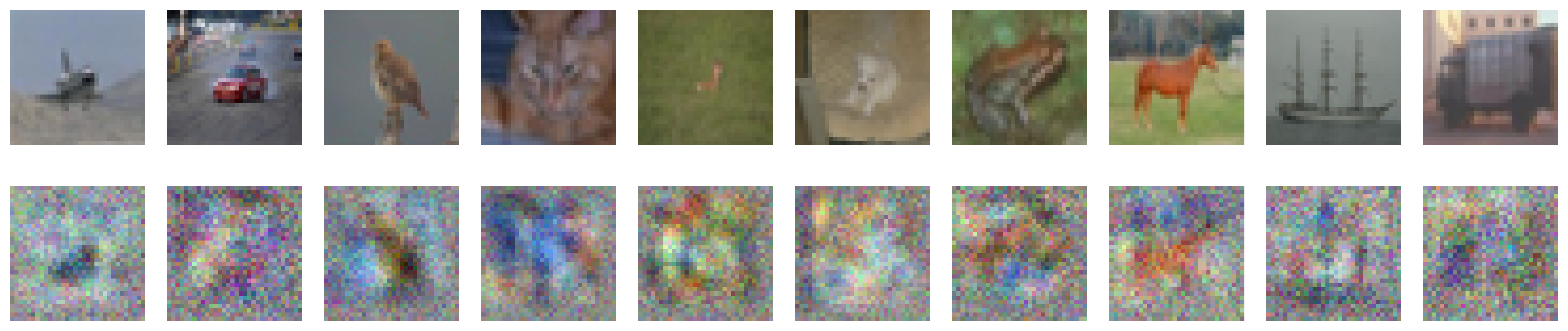} 
    \caption{The first row shows class embeddings initialized using the image with the smallest median distance to all other images within the same class for CIFAR-10. The second row displays the learned class embeddings from NoProp-DT, which can be interpreted as learned image prototypes for each class.}
    \label{fig:prototypes}
\end{figure}

\begin{figure}[htbp]
    \centering
    \begin{minipage}{0.32\textwidth} 
        \centering
        \includegraphics[width=\linewidth]{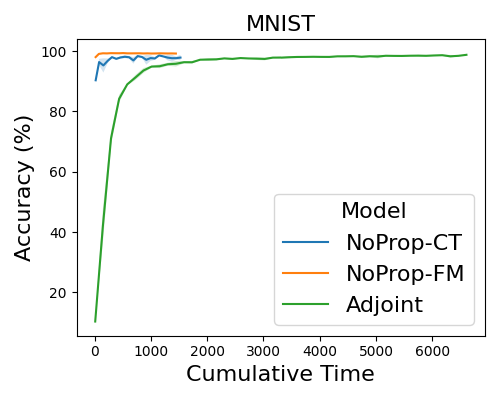} 
    \end{minipage}%
    \begin{minipage}{0.32\textwidth} 
        \centering
        \includegraphics[width=\linewidth]{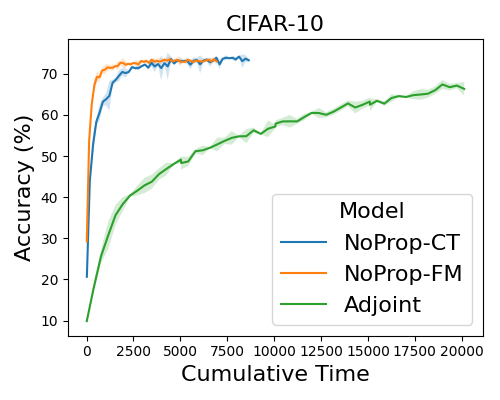} 
    \end{minipage}%
    \begin{minipage}{0.32\textwidth} 
        \centering
        \includegraphics[width=\linewidth]{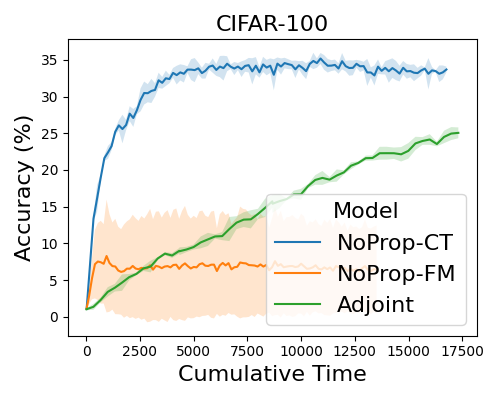} 
    \end{minipage}
    \caption{Test accuracies (\%) plotted against cumulative training time (in seconds) for models using one-hot label embedding in the continuous-time setting. All models within each plot were trained on the same type of GPU to ensure a fair comparison. NoProp-CT achieves strong performance in terms of both accuracy and speed compared to adjoint sensitivity. For CIFAR-100, NoProp-FM does not learn effectively with one-hot label embedding.} 
    \label{fig:continuous}
\end{figure}

\begin{table}
    \centering
    \begin{tabular}{llll}
    \hline
    Method    & MNIST   & CIFAR-10 & CIFAR-100 \\ \hline
    \multicolumn{4}{c}{Discrete-time}          \\ \hline
    Backprop  & 0.87 GB & 1.17 GB  & 1.73 GB   \\
    NoProp-DT & 0.49 GB & 0.64 GB  & 1.23 GB   \\ \hline
    \multicolumn{4}{c}{Continuous-time}        \\ \hline
    Adjoint   & 2.32 GB & 6.23 GB  & 6.45 GB   \\
    NoProp-CT & 1.05 GB & 0.45 GB  & 0.50 GB   \\
    NoProp-FM & 1.06 GB & 0.44 GB  & 0.49 GB   \\ \hline
    \end{tabular}
    \caption{Process GPU memory allocated (in GB) for models using one-hot label embedding.}
    \label{table:memory}
\end{table}


\paragraph{Main results}
Our main results, summarised in Table~\ref{table:results}, demonstrate that NoProp-DT achieves performance comparable to or better than back-propagation on MNIST, CIFAR-10, and
CIFAR-100 in the discrete-time setting. We also report results from prior alternative methods to full back-propagation, including the Forward-Forward Algorithm, Difference Target Propagation \citep{lee2015difference}, and the Local Greedy Forward Gradient Activity-Perturbed method \citep{ren2022scaling}. However, direct comparisons are challenging because NoProp employs a fundamentally different framework based on conditional diffusion or flow matching, where the input is broadcast to each block (see Figure~\ref{fig:backprop_alternatives}). While our results surpass previously reported performances of these methods, differences in architectures and model sizes make direct comparisons difficult, so improvements cannot be solely credited to our approach. Nevertheless, our results demonstrate that NoProp can achieve non-trivial performance without requiring full forward and backward passes through the entire network. We also conducted a limited evaluation on Tiny ImageNet, comparing NoProp-DT with back-propagation, with results shown in Table~\ref{table:results_tiny_imagenet}. While the method has not been fully optimized in this paper and should be viewed as a proof of concept, it provides preliminary evidence that NoProp has the potential to scale beyond small datasets. Additionally, NoProp demonstrates reduced GPU memory consumption during training, as shown in Table~\ref{table:memory}. To illustrate the learned class embeddings, Figure~\ref{fig:prototypes} visualises both the initializations and the final class embeddings for CIFAR-10 learned by NoProp-DT, where the embedding dimension matches the image dimension. 

In the continuous setting, NoProp-CT and NoProp-FM achieve lower accuracy than NoProp-DT, likely due to the additional conditioning on time \( t \). However, they generally outperform the adjoint sensitivity method on CIFAR-10 and CIFAR-100, both in terms of accuracy and computational efficiency. While the adjoint method achieves similar accuracy to NoProp-CT and NoProp-FM on MNIST, it does so much slower, as shown in Figure~\ref{fig:continuous}. 

For CIFAR-100 with one-hot embeddings, NoProp-FM fails to learn effectively, resulting in very slow accuracy improvement. In contrast, NoProp-CT still outperforms the adjoint method. However, once label embeddings are learned jointly, the performance of NoProp-FM improves significantly.

We also conducted ablation studies on the parameterisations of class probabilities, $\hat{p}_{\theta_{\textrm{out}}}(y|z_T)$, and the initializations of the class embedding matrix, $W_{\text{Embed}}$, with results shown in Figure~\ref{fig:class_probabilities} and Figure~\ref{fig:initializations} in the Appendix. The ablation results reveal no consistent advantage between the class probability parameterisations, with performance varying across datasets. For class embedding initializations, random initializations are comparable to orthogonal and prototype initializations.

\section{Conclusion}
Using the denoising score matching approach that underlies diffusion models, we have proposed NoProp, a novel approach for training neural networks without full forward- or back-propagation across network blocks. The method enables each block of the neural network to be trained independently to predict the target label given a noisy label and the training input, while at inference time, each block takes the noisy label produced by the previous block and denoises it by taking a step towards the label it predicts. Our experiments show that NoProp is a viable learning algorithm, achieving performance comparable to full back-propagation on the same architecture, while offering advantages in biological plausibility and enabling parallel training. However, our experiments focus on relatively small-scale classification datasets, and evaluating NoProp on larger datasets with higher-resolution inputs will be an important direction for future work. In particular, scaling will require addressing the cost of broadcasting high-dimensional inputs to all blocks and potentially optimizing communication strategies between blocks while preserving locality. While we have applied NoProp only to classification tasks, the underlying framework of block-wise local training may be useful in broader settings. For example, in continual learning, the lack of gradient flow across blocks might help localize task interference, potentially reducing catastrophic forgetting. Although we do not investigate this directly, we highlight it as a possible direction for future research. We believe that the perspective of training neural networks via denoising score matching opens up new possibilities for training deep learning models without full back-propagation, and we hope that our work will inspire further research in this direction.

\section*{Acknowledgments}
Qinyu Li is supported by the Oxford-Radcliffe Graduate Scholarship and the EPSRC CDT in Modern Statistics and Statistical Machine Learning (EP/S023151/1).

\newpage

\bibliography{references.bib}
\bibliographystyle{collas2025_conference}

\newpage
\appendix

\section{Derivations of training objectives of NoProp} \label{sec:derivations}
For completeness, we include derivations of the training objectives of NoProp-DT and NoProp-CT, closely following \citet{sohlDickstein15} and \citet{kingma2021variational}. 

\subsection{Derivation of Equation~\ref{eq:elbo}} \label{sec:elbo}
\begin{align}
    \log p(y|x) &= \log \int p((z_t)_{t=0}^T,y|x)d(z_t)_{t=0}^T \\
    &= \log \int \frac{p((z_t)_{t=0}^T,y|x) q((z_t)_{t=0}^T|y,x)}{q((z_t)_{t=0}^T|y,x)} d(z_t)_{t=0}^T \\
    &= \log \E_{q((z_t)_{t=0}^T|y,x)} \left[ \frac{p((z_t)_{t=0}^T,y|x)}{q((z_t)_{t=0}^T|y,x)} \right] \\
    &\ge \E_{q((z_t)_{t=0}^T|y,x)} \left[ \log p((z_t)_{t=0}^T,y|x) - \log q((z_t)_{t=0}^T|y,x) \right].
\end{align}
The last step follows from Jensen's inequality, yielding a lower bound on $\log p(y | x)$. This bound is commonly referred to as the evidence lower bound (ELBO).

\subsection{$q(z_{t-1}|z_{t})$ and $q(z_t|y)$} \label{sec:diffusion_noising}
We will use the reparameterization trick to derive the expressions for $q(z_t|y)$ and $q(z_t|z_s)$. The reparameterization trick allows us to rewrite a random variable sampled from a distribution in terms of a deterministic function of a noise variable. Specifically, for a Gaussian random variable $z \sim \N(z|\mu, \sigma^2)$, we can reparameterize it as:
\begin{equation}
    z = \mu + \sigma \epsilon, \quad \epsilon \sim \N(\epsilon | 0,1).
\end{equation}
Let $\{\epsilon_t^*, \epsilon_t\}_{t=0}^T, \epsilon_y^*, \epsilon_y \sim \N(0, 1)$. Then, for any sample $z_t \sim q(z_t | z_s)$ for any $0 \le t < s \le T$, it can be expressed as
\begin{align}
    z_t &= \sqrt{\alpha_t} z_{t+1} + \sqrt{1 - \alpha_t} \epsilon_{t+1}^* \\
    &= \sqrt{\alpha_t}(\sqrt{\alpha_{t+1}}z_{t+2} + \sqrt{1 - \alpha_{t+1}}\epsilon_{t+2}^*) + \sqrt{1 - \alpha_t} \epsilon_{t+1}^*\\
    &= \sqrt{\alpha_t \alpha_{t+1}} z_{t+2} + \sqrt{\alpha_t - \alpha_t \alpha_{t+1}} \epsilon_{t+2}^* + \sqrt{1 - \alpha_t} \epsilon_{t+1}^* \\
    &= \sqrt{\alpha_t \alpha_{t+1}} z_{t+2} + \sqrt{(\alpha_t - \alpha_t \alpha_{t+1}) + (1 - \alpha_t)} \epsilon_{t+2} \\ 
    &= \sqrt{\alpha_t \alpha_{t+1}} z_{t+2} + \sqrt{1 - \alpha_t \alpha_{t+1}} \epsilon_{t+2} \\
    &= \sqrt{\prod_{i=t}^{s-1} \alpha_i} z_s + \sqrt{1 - \prod_{i=t}^{s-1} \alpha_i} \epsilon_s \\
    &= \sqrt{\frac{\bar{\alpha}_t}{\bar{\alpha}_s}} z_s + \sqrt{1 - \frac{\bar{\alpha}_t}{\bar{\alpha}_s}} \epsilon_s.
\end{align}
Hence, 
\begin{equation}
    q(z_t | z_s) = \N (z_t | \sqrt{\frac{\bar{\alpha}_t}{\bar{\alpha}_s}}, 1 - \frac{\bar{\alpha}_t}{\bar{\alpha}_s}).
\end{equation}
In particular, 
\begin{equation}
    q(z_{t-1} | z_t) = \N (z_{t-1} | \sqrt{\alpha_{t-1}} z_t, 1 - \alpha_{t-1}).
\end{equation} 
Similarly, it can be shown that 
\begin{equation}
    q(z_t|y) = \N(z_t|\sqrt{\baralpha_t}u_y,1-\baralpha_t). 
\end{equation}

\subsection{$q(z_t | z_{t-1}, y)$} \label{sec:diffusion_denoising}
Applying the Bayes' theorem, we can obtain the posterior density 
\begin{align}
    q(z_t | z_{t-1}, y) &\propto q(z_t | y) q(z_{t-1} | z_t) \\
    &\propto \N(z_t | \sqrt{\baralpha_t} u_y, 1 - \baralpha_t) \N(z_{t-1} | \sqrt{\baralpha_{t-1}} z_t, 1 - \alpha_{t-1}) \\
    &\propto \exp \left( -\frac{1}{2(1 - \baralpha_t)} (z_t - \sqrt{\baralpha_t} u_y)^T (z_t - \sqrt{\baralpha_t} u_y) \right.\\
    \quad & \left. - \frac{1}{2(1 - \alpha_{t-1})} (z_{t-1} - \sqrt{\alpha_{t-1}}z_t)^T(z_{t-1} - \sqrt{\alpha_{t-1}}z_t)\right) \\
    &\propto \exp \left( -\frac{1}{2c_t} (z_t - \mu_t(z_{t-1}, u_y))^T(z_t - \mu_t(z_{t-1}, u_y)) \right).
\end{align}
Hence we have $q(z_t | z_{t-1}, y) = \N(z_t | \mu_t(z_{t-1}, u_y), c_t)$, where 
\begin{align}
    \mu_t(z_{t-1},u_y) &= \frac{\sqrt{\baralpha_{t}}(1-\alpha_{t-1})}{1-\baralpha_{t-1}} u_y 
    + \frac{\sqrt{\alpha_{t-1}}(1-\baralpha_{t})}{1-\baralpha_{t-1}} z_{t-1}, \\
    c_t &= \frac{(1-\baralpha_{t})(1-\alpha_{t-1})}{1-\baralpha_{t-1}}.
\end{align}

\subsection{Objectives of NoProp-DT and NoProp-CT} \label{sec:objectives}

Starting from the ELBO derived in Appendix~\ref{sec:elbo}, we have 
\begin{align}
    \log p(y|x) & \ge \E_{q((z_t)_{t=0}^T|y,x)} \left[ \log \frac{p((z_t)_{t=0}^T,y|x)}{q((z_t)_{t=0}^T|y, x)} \right] \\
    &= \E_{q((z_t)_{t=0}^T|y)} \left[ \log \frac{p(z_0)\left(\prod_{t=1}^{T} p(z_{t}|z_{t-1},x)\right) p(y|z_T)}{q(z_T|y)\left(\prod_{t=T}^1 q(z_{t-1}|z_{t})\right)} \right] \\
    &= \E_{q((z_t)_{t=0}^T|y)} \left[ \log \frac{p(z_0) p(y|z_T)}{q(z_T|y)} + \log \prod_{t=1}^T \frac{p(z_{t}|z_{t-1},x)}{q(z_{t-1}|z_{t}, y)}\right] \\
    &= \E_{q((z_t)_{t=0}^T|y)} \left[ \log \frac{p(z_0) p(y|z_T)}{q(z_T|y)} + \log \prod_{t=1}^T \frac{p(z_{t}|z_{t-1},x)}{\frac{q(z_t|z_{t-1}, y)q(z_{t-1}|y)}{q(z_t|y)}}\right] \\
    &= \E_{q((z_t)_{t=0}^T|y)} \left[ \log \frac{p(z_0) p(y|z_T)}{q(z_T|y)} + \log \frac{q(z_T|y)}{q(z_0|y)} + \log \prod_{t=1}^T \frac{p(z_{t}|z_{t-1},x)}{q(z_t|z_{t-1}, y)}\right] \\
    &= \E_{q((z_t)_{t=0}^T|y)} \left[ \log \frac{p(z_0) p(y|z_T)}{q(z_0|y)} + \log \prod_{t=1}^T \frac{p(z_{t}|z_{t-1},x)}{q(z_t|z_{t-1}, y)}\right] \\
    &= \E_{q((z_t)_{t=0}^T|y)} \left[ \log p(y|z_T) \right] + \E_{q((z_t)_{t=0}^T|y)} \left[ \log \frac{p(z_0)}{q(z_0|y)}  \right] + \sum_{t=1}^T \E_{q((z_t)_{t=0}^T|y)} \left[ \log \frac{p(z_{t}|z_{t-1},x)}{q(z_t|z_{t-1}, y)} \right] \\
    &= \E_{q(z_T|y)} \left[ \log p(y|z_T) \right] + \E_{q(z_0|y)} \left[ \log \frac{p(z_0)}{q(z_0|y)}  \right] + \sum_{t=1}^T \E_{q(z_{t-1}, z_t|y)} \left[ \log \frac{p(z_{t}|z_{t-1},x)}{q(z_t|z_{t-1}, y)} \right] \\
    &= \E_{q(z_T|y)} \left[ \log p(y|z_T) \right] - D_{\mathrm{KL}}(q(z_0 | y) \Vert p(z_0)) - \sum_{t=1}^T \E_{q(z_{t-1}|y)} \left[D_{\mathrm{KL}}(q(z_t | z_{t-1}, y) \Vert p(z_{t}|z_{t-1},x))\right].
\end{align}
Since $q(z_t | z_{t-1}, y)$ and $p(z_{t}|z_{t-1},x)$ are both Gaussian, with $q(z_t | z_{t-1}, y) = \N(z_t | \mu_t(z_{t-1}, u_y), c_t)$ and $p(z_{t}|z_{t-1},x) = \N(z_t | \mu_t(z_{t-1}, \hatu_{\theta_t}(z_{t-1},x)), c_t)$, their KL divergence is available in closed form:
\begin{align}
    D_{\mathrm{KL}}(q(z_t | z_{t-1}, y) \Vert p(z_{t}|z_{t-1},x)) &= \frac{1}{2c_t} \| \mu_t(z_{t-1}, \hatu_{\theta_t}(z_{t-1},x)) -  \mu_t(z_{t-1}, u_y) \|^2 \\
    &= \frac{1}{2c_t} \frac{\baralpha_t(1 - \alpha_{t-1})^2}{(1 - \baralpha_{t-1})^2} \| \hatu_{\theta_t}(z_{t-1},x) - u_y \|^2 \\
    &= \frac{1 - \baralpha_{t-1}}{2(1 - \baralpha_t)(1 - \alpha_{t-1})} \frac{\baralpha_t(1 - \alpha_{t-1})^2}{(1 - \baralpha_{t-1})^2} \| \hatu_{\theta_t}(z_{t-1},x) - u_y \|^2 \\
    &= \frac{\baralpha_t(1 - \alpha_{t-1})}{2(1 - \baralpha_t)(1 - \baralpha_{t-1})} \| \hatu_{\theta_t}(z_{t-1},x) - u_y \|^2 \\
    &= \frac{1}{2}(\frac{\baralpha_t}{1 - \baralpha_t} - \frac{\baralpha_{t-1}}{1 - \baralpha_{t-1}}) \| \hatu_{\theta_t}(z_{t-1},x) - u_y \|^2 \\
    &= \frac{1}{2} (\textrm{SNR}(t) - \textrm{SNR}(t-1)) \| \hatu_{\theta_t}(z_{t-1},x) - u_y \|^2.
\end{align}
Using $\hat{p}_{\theta_{\textrm{out}}}(y|z_T)$ to estimate $p(y|z_T)$, the ELBO becomes 
\begin{equation}
    \E_{q(z_T|y)} \left[ \log \hat{p}_{\theta_{\textrm{out}}}(y|z_T) \right] - D_{\mathrm{KL}}(q(z_0 | y) \Vert p(z_0)) - \frac{1}{2} \sum_{t=1}^T \E_{q(z_{t-1}|y)} \left[(\textrm{SNR}(t) - \textrm{SNR}(t-1)) \| \hatu_{\theta_t}(z_{t-1},x) - u_y \|^2 \right].
\end{equation}
Instead of computing all $T$ terms in the sum, we replace it with an unbiased estimator, which gives 
\begin{equation}
    \E_{q(z_T|y)} \left[ \log \hat{p}_{\theta_{\textrm{out}}}(y|z_T) \right] - D_{\mathrm{KL}}(q(z_0 | y) \Vert p(z_0)) - \frac{T}{2} \E_{t \sim \mathcal{U}\{1, T\}, q(z_{t-1}|y)} \left[(\textrm{SNR}(t) - \textrm{SNR}(t-1)) \| \hatu_{\theta_t}(z_{t-1},x) - u_y \|^2 \right].
\end{equation}
With an additional hyperparameter $\eta$ in Equation~\ref{eq:noprop_dt}, this yields the NoProp objective in the discrete-time case. However, note that in our experiments, we choose to iterate over the values of $t$ from 1 to $T$ instead of sampling random values. Details can be found in Algorithm~\ref{alg:noprop_dt}. 

In the continuous-time case where $t$ is scaled to the range (0, 1), let $\tau = 1/T$. If we rewrite the term involving SNR in the NoProp objective as
\begin{equation}
    \frac{1}{2} \E_{t \sim \mathcal{U}[0, 1]} \left[\frac{\textrm{SNR}(t + \tau) - \textrm{SNR}(t)}{\tau} \| \hatu_{\theta}(z_{t},x, t) - u_y \|^2 \right].
\end{equation}
As $T \to \infty$, this becomes 
\begin{equation}
    \frac{1}{2} \E_{t \sim \mathcal{U}\{0, 1\}} \left[\textrm{SNR}'(t) \| \hatu_{\theta}(z_{t},x, t) - u_y \|^2 \right].
\end{equation}
Again, with an additional hyperparameter $\eta$, this gives the NoProp objective in the continuous-time case in Equation~\ref{eq:noprop_ct}. 

\begin{algorithm}[h!]
    \caption{\textbf{NoProp-DT (Training)}}
    \label{alg:noprop_dt}
    \begin{algorithmic}
        \Require $T$ diffusion steps, dataset $\{(x_i, y_i)\}_{i=1}^{N}$, batch size $B$, hyperparameter $\eta$, embedding matrix $W_{\text{Embed}}$, parameters $\{\theta_t\}_{t=1}^{T}, \theta_{\text{out}}$, noise schedule $\{\alpha_t\}_{t=0}^{T}$
        \For{$t = 1$ to $T$}
            \For{each mini-batch $\mathcal{B} \subset \{(x_i, y_i)\}_{i=1}^{N}$ of size $B$}
                \For{each $(x_i, y_i) \in \mathcal{B}$}
                    \State Obtain label embedding $u_{y_i} = \{W_{\text{Embed}}\}_{y_i}$.
                    \State Sample $z_{t,i} \sim \N(z_{t, i}|\sqrt{\baralpha_t}u_{y, i},1-\baralpha_t)$.
                \EndFor
                \State Compute the loss function:
                \begin{align}
                    \mathcal{L}_{t} &= \frac{1}{B} \sum_{i \in \mathcal{B}} \left[ -\log \hat{p}_{\theta_{\text{out}}}(y_i | z_{T,i}) \right] \nonumber \\
                    &\quad + \frac{1}{B} \sum_{i \in \mathcal{B}} D_{\mathrm{KL}}(q(z_{0} | y_i) \Vert p(z_{0})) \nonumber \\
                    &\quad + \frac{T}{2B} \eta \sum_{i \in \mathcal{B}} \left( \text{SNR}(t) - \text{SNR}(t-1) \right) \left\| \hat{u}_{\theta_t}(z_{t-1,i}, x_i) - u_{y_i} \right\|^2.
                \end{align}
                \State Update $\theta_t$, $\theta_{\text{out}}$, and $W_{\text{Embed}}$ using gradient-based optimization.
            \EndFor
        \EndFor
    \end{algorithmic}
\end{algorithm}

\begin{algorithm}[h]
    \caption{\textbf{NoProp-CT (Training)}}
    \label{alg:noprop_ct}
    \begin{algorithmic}
        \Require dataset $\{(x_i, y_i)\}_{i=1}^{N}$, batch size $B$, hyperparameter $\eta$, embedding matrix $W_{\text{Embed}}$, parameters $\theta, \theta_{\text{out}}$, noise schedule $\bar{\alpha}_t = \sigma(-\gamma_\psi(t))$
        \For{each mini-batch $\mathcal{B} \subset \{(x_i, y_i)\}_{i=1}^{N}$ with size $B$}
            \For{each $(x_i, y_i) \in \mathcal{B}$}
                \State Obtain label embedding $u_{y_i} = \{W_{\text{Embed}}\}_{y_i}$.
                \State Sample $t_i \sim \mathcal{U}(0,1)$.
                \State Sample $z_{t_i,i} \sim \N(z_{t_i, i}|\sqrt{\bar{\alpha}_{t_i}}u_{y, i},1-\bar{\alpha}_{t_i})$.
            \EndFor
            \State Compute the loss function:
            \begin{align}
                \mathcal{L} &= \frac{1}{B} \sum_{i \in \mathcal{B}} \left[ -\log \hat{p}_{\theta_{\text{out}}}(y_i | z_{1,i}) \right] \nonumber \\
                &\quad + \frac{1}{B} \sum_{i \in \mathcal{B}} D_{\mathrm{KL}}(q(z_{0} | y_i) \Vert p(z_{0})) \nonumber \\
                &\quad + \frac{1}{2B} \eta \sum_{i \in \mathcal{B}} \textrm{SNR}'(t_i) \left\| \hat{u}_{\theta}(z_{t_i,i}, x_i, t_i) - u_{y_i} \right\|^2.
            \end{align}
            \State Update $\theta$, $\theta_{\text{out}}$, $\psi$, and $W_{\text{Embed}}$ using gradient-based optimization.
        \EndFor
    \end{algorithmic}
\end{algorithm}

\begin{algorithm}[h]
    \caption{\textbf{NoProp-FM (Training)}}
    \label{alg:noprop_fm}
    \begin{algorithmic}
        \Require dataset $\{(x_i, y_i)\}_{i=1}^{N}$, batch size $B$, embedding matrix $W_{\text{Embed}}$, parameters $\theta, \theta_{\text{out}}$
        \For{each mini-batch $\mathcal{B} \subset \{(x_i, y_i)\}_{i=1}^{N}$ with size $B$}
            \For{each $(x_i, y_i) \in \mathcal{B}$}
                \State Obtain label embedding $u_{y_i} = \{W_{\text{Embed}}\}_{y_i}$ and set $z_{1, i} = u_{y_i}$.
                \State Sample $z_{0, i} \sim \N(z_{0, i} | 0, 1)$.
                \State Sample $t_i \sim \mathcal{U}(0,1)$.
                \State Sample $z_{t_i, i} \sim \N(z_{t_i, i} \mid t_i z_{1, i} + (1-t_i) z_{0, i}, \sigma^2)$.
            \EndFor
            \State Compute the loss function:
            \begin{equation}
                \mathcal{L} = \frac{1}{B} \sum_{i \in \mathcal{B}} \lVert v_{\theta}(z_{t_i, i}, x_i, t_i) - (z_{1, i} - z_{0, i}) \rVert^2.
            \end{equation}
            \If{$W_{\text{Embed}}$ has learnable parameters}
                \For{each $(x_i, y_i) \in \mathcal{B}$}
                    \State Compute the extrapolated linear estimate $\tilde{z}_{1, i} = z_{t_i, i} + (1-t_i) v_{\theta}(z_{t_i, i}, x_i, t_i)$.
                \EndFor
                \State Modify the loss function:
                \begin{equation}
                    \mathcal{L} \gets \mathcal{L} - \frac{1}{B} \sum_{i \in \mathcal{B}} \log \hat{p}_{\theta_{\mathrm{out}}}(y_i \mid \tilde{z}_{1, i}).
                \end{equation}
            \EndIf
            \State Update $\theta$, $\theta_{\text{out}}$, and $W_{\text{Embed}}$ using gradient-based optimization.
        \EndFor
    \end{algorithmic}
\end{algorithm}

\section{Trainable noise schedule for NoProp-CT} \label{sec:noise_schedule}
Following \citet{kingma2021variational} and \citet{gulrajani2024likelihood}, we parameterize the signal-to-noise ratio (SNR) as  
\begin{equation}
    \text{SNR}(t) = \exp(-\gamma(t)),
\end{equation}
where \(\gamma(t)\) is a learnable function that determines the rate of noise decay. To ensure consistency with our formulation, \(\gamma(t)\) must be monotonically decreasing in \(t\). 

We implement \(\gamma(t)\) using a neural network-based parameterization. Specifically, we define an intermediate function \(\bar{\gamma}(t)\), which is normalized to the unit interval:
\begin{equation}
    \bar{\gamma}(t) = \frac{\tilde{\gamma}(t) - \tilde{\gamma}(0)}{\tilde{\gamma}(1) - \tilde{\gamma}(0)},
\end{equation}
where \(\tilde{\gamma}(t)\) is modeled as a two-layer neural network where the weights are restricted to be positive. 

To align with our paper's formulation where \(\gamma(t)\) should decrease with \(t\), we define:
\begin{equation}
    \gamma(t) = \gamma_0 + (\gamma_1 - \gamma_0) (1 - \bar{\gamma}(t)),
\end{equation}
where \(\gamma_0\) and \(\gamma_1\) are trainable endpoints of the noise schedule. Finally, we obtain the noise schedule $\baralpha_t = \sigmoid(-\gamma(t))$.


\begin{table}[]
    \centering
    \begin{tabular}{@{}lllllllll@{}}
    \toprule
    Dataset                    & Method    & Batch Size & Epochs & Optimiser & Learning Rate & Weight Decay & Timesteps & $\eta$ \\ \midrule
    \multirow{5}{*}{MNIST}     & Backprop  & 128        & 100    & AdamW     & 0.001         & 0.001        & 10        & -      \\
                               & NoProp-DT & 128        & 100    & AdamW     & 0.001         & 0.001        & 10        & 0.1     \\
                               & Adjoint   & 128        & 2      & AdamW     & 0.001         & 0.001        & 1000      & -      \\
                               & NoProp-CT & 128        & 100    & Adam      & 0.001         & 0.001        & 1000      & 1      \\
                               & NoProp-FM & 128        & 100    & Adam      & 0.001         & 0.001        & 1000      & -      \\ \midrule
    \multirow{5}{*}{CIFAR-10}  & Backprop  & 128        & 150    & AdamW     & 0.001         & 0.001        & 10        & -      \\
                               & NoProp-DT & 128        & 150    & AdamW     & 0.001         & 0.001        & 10        & 0.1     \\
                               & Adjoint   & 128        & 4      & AdamW     & 0.001         & 0.001        & 1000      & -      \\
                               & NoProp-CT & 128        & 500    & Adam      & 0.001         & 0.001        & 1000      & 1      \\
                               & NoProp-FM & 128        & 500    & Adam      & 0.001         & 0.001        & 1000      & -      \\ \midrule
    \multirow{5}{*}{CIFAR-100} & Backprop  & 128        & 150    & AdamW     & 0.001         & 0.001        & 10        & -      \\
                               & NoProp-DT & 128        & 150    & AdamW     & 0.001         & 0.001        & 10        & 0.1     \\
                               & Adjoint   & 128        & 4      & AdamW     & 0.001         & 0.001        & 1000      & -      \\
                               & NoProp-CT & 128        & 1000   & Adam      & 0.001         & 0.001        & 1000      & 1      \\
                               & NoProp-FM & 128        & 1000   & Adam      & 0.001         & 0.001        & 1000      & -      \\ \bottomrule
    \end{tabular}
    \caption{Experiment details. $\eta$ is the hyperparameter in Equations~\ref{eq:noprop_dt} and \ref{eq:noprop_ct}.}
    \label{table:experiment_details}
    \end{table}

    \begin{figure}[htbp]
        \centering
        \begin{minipage}{0.3\textwidth} 
            \centering
            \includegraphics[width=\linewidth]{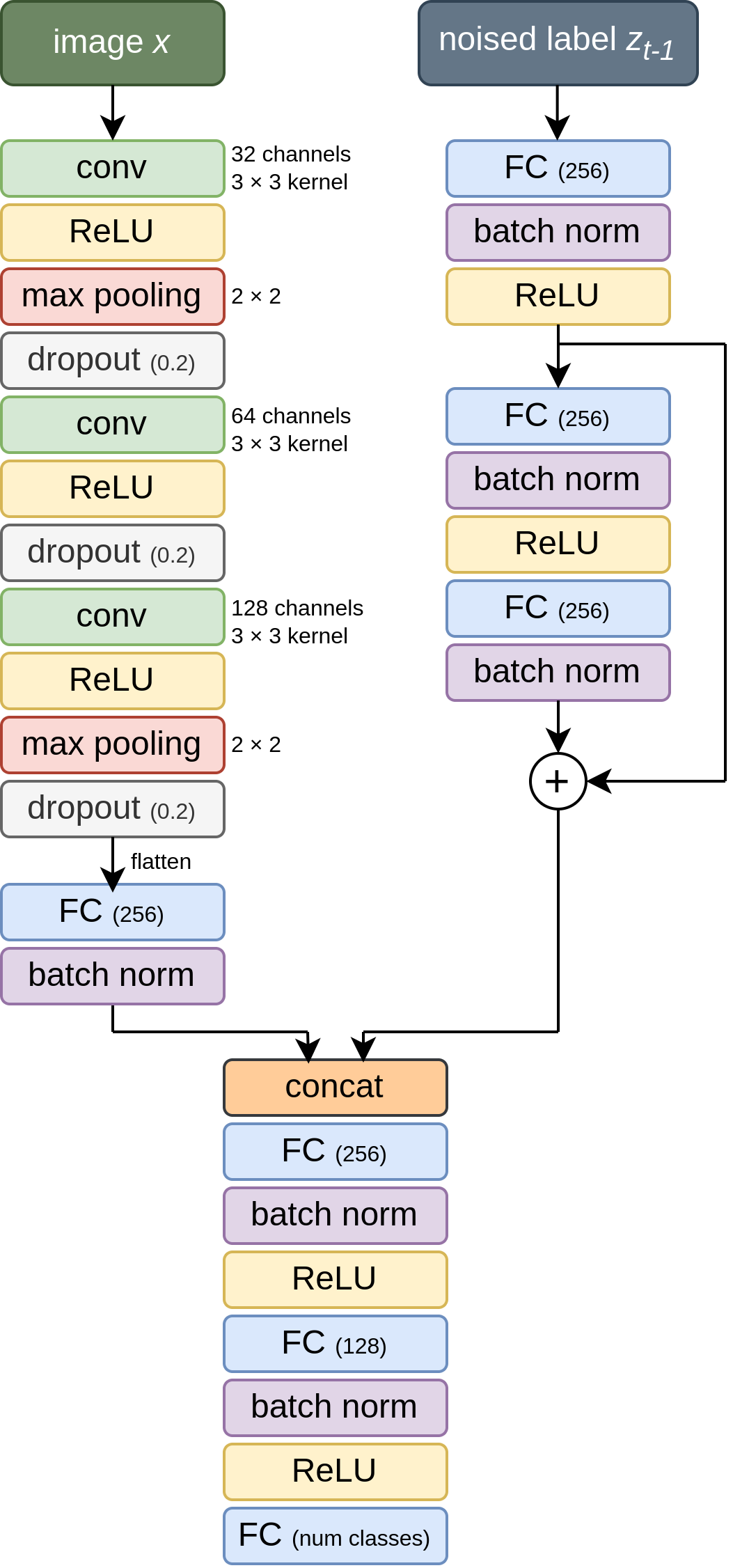} 
        \end{minipage}%
        \hspace{0.2\textwidth} 
        \begin{minipage}{0.4\textwidth} 
            \centering
            \includegraphics[width=\linewidth]{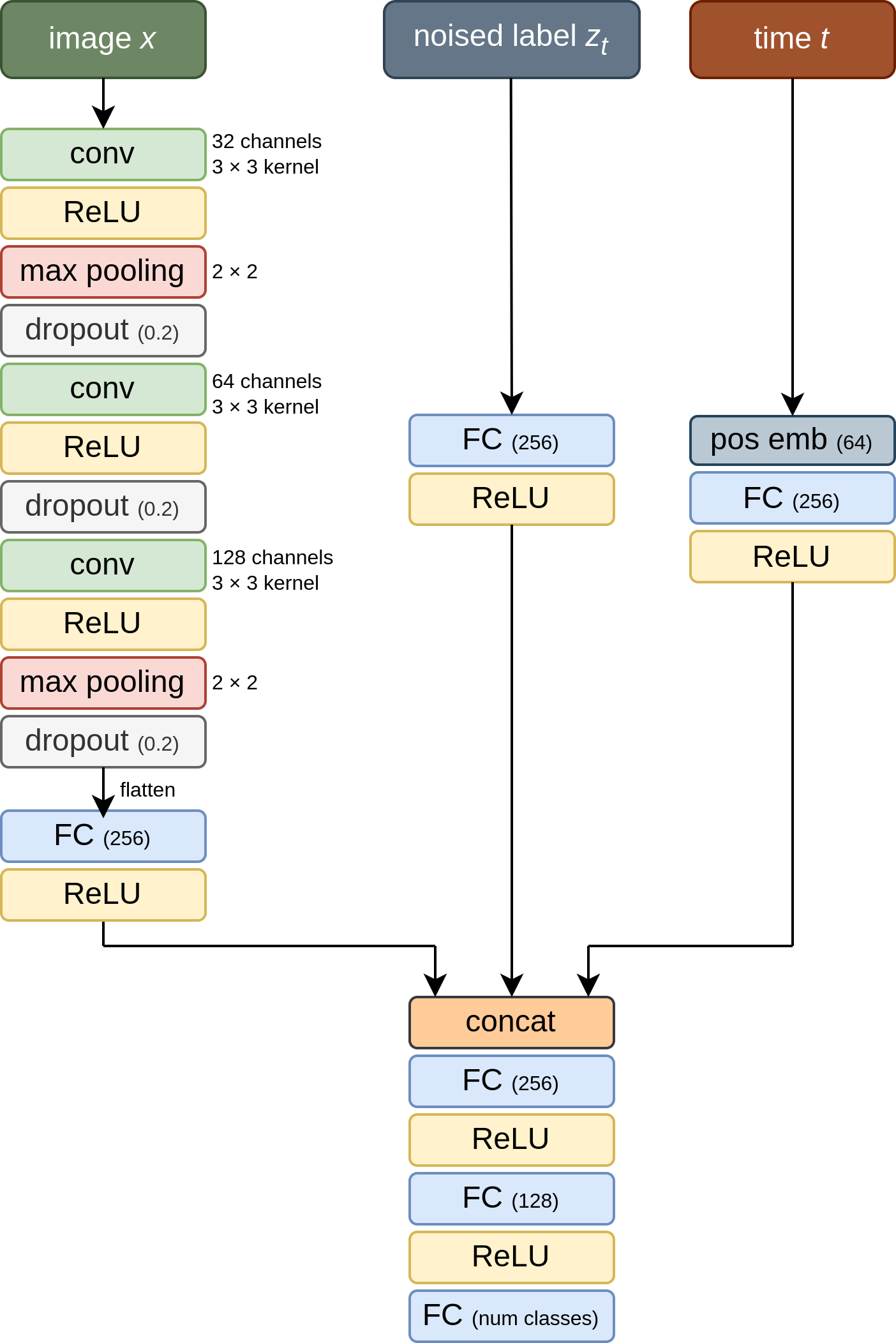} 
        \end{minipage}%
        \caption{Models used for training when the class embedding dimension is different from the image dimension. Left: model for the discrete-time case. For Tiny ImageNet, an additional convolutional layer with 256 output channels, followed by ReLU, 2×2 max pooling, and dropout (p = 0.2), is inserted before concatenation. Right: model for the continuous-time case. conv: convolutional layer. FC: fully connected layer (number in parentheses indicates units). concat: concatenation. pos emb: positional embedding (number in parentheses indicates time embedding dimension). When the class embedding dimension matches the image dimension, the noised label and the image are processed in the same way before concatenation in each model. Note that batch normalization is not included in the continuous-time model.} 
        \label{fig:models}
    \end{figure}

    \begin{figure}[htbp]
        \centering
        \begin{minipage}{0.31\textwidth} 
            \centering
            \includegraphics[width=\linewidth]{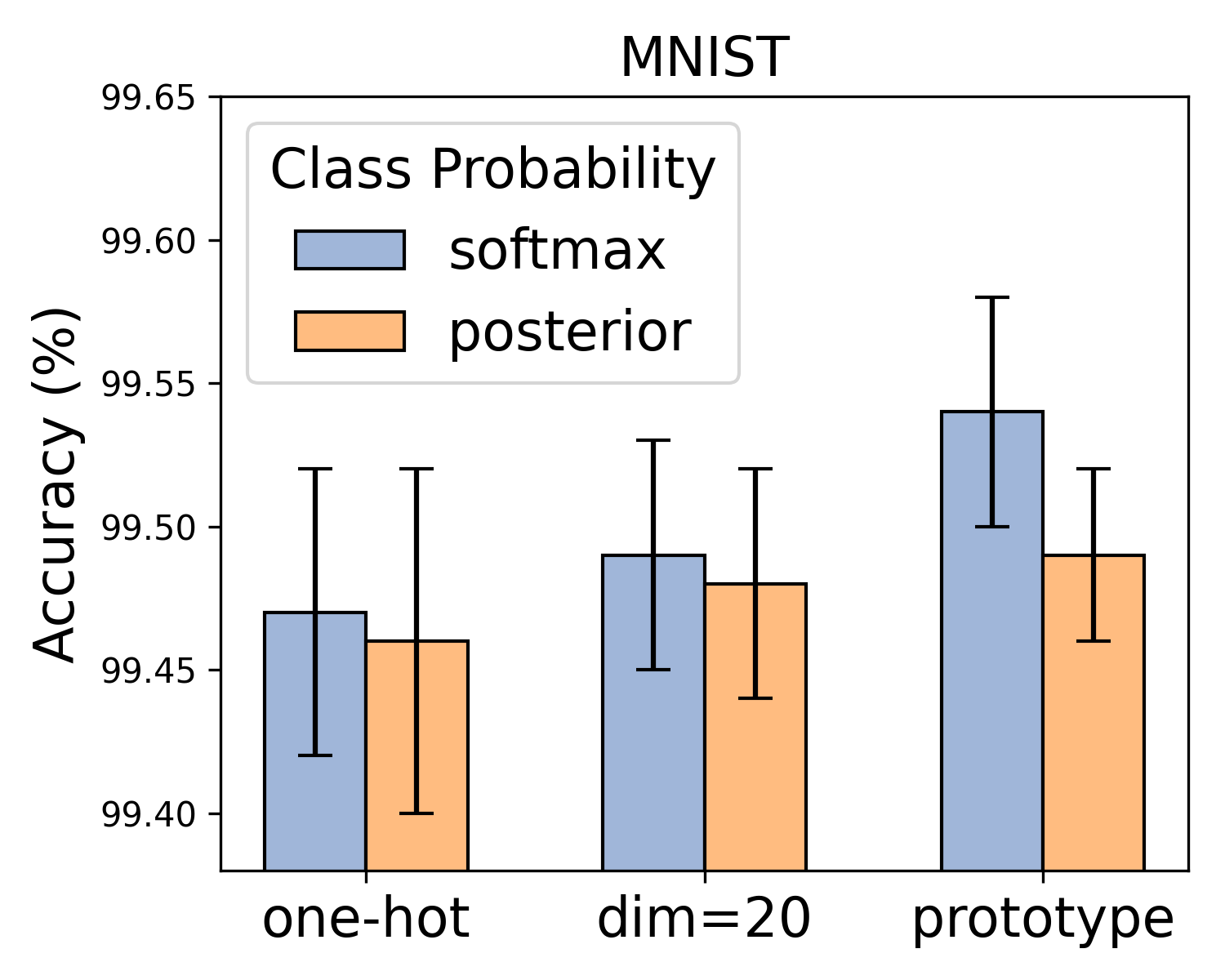} 
        \end{minipage}%
        \begin{minipage}{0.31\textwidth} 
            \centering
            \includegraphics[width=\linewidth]{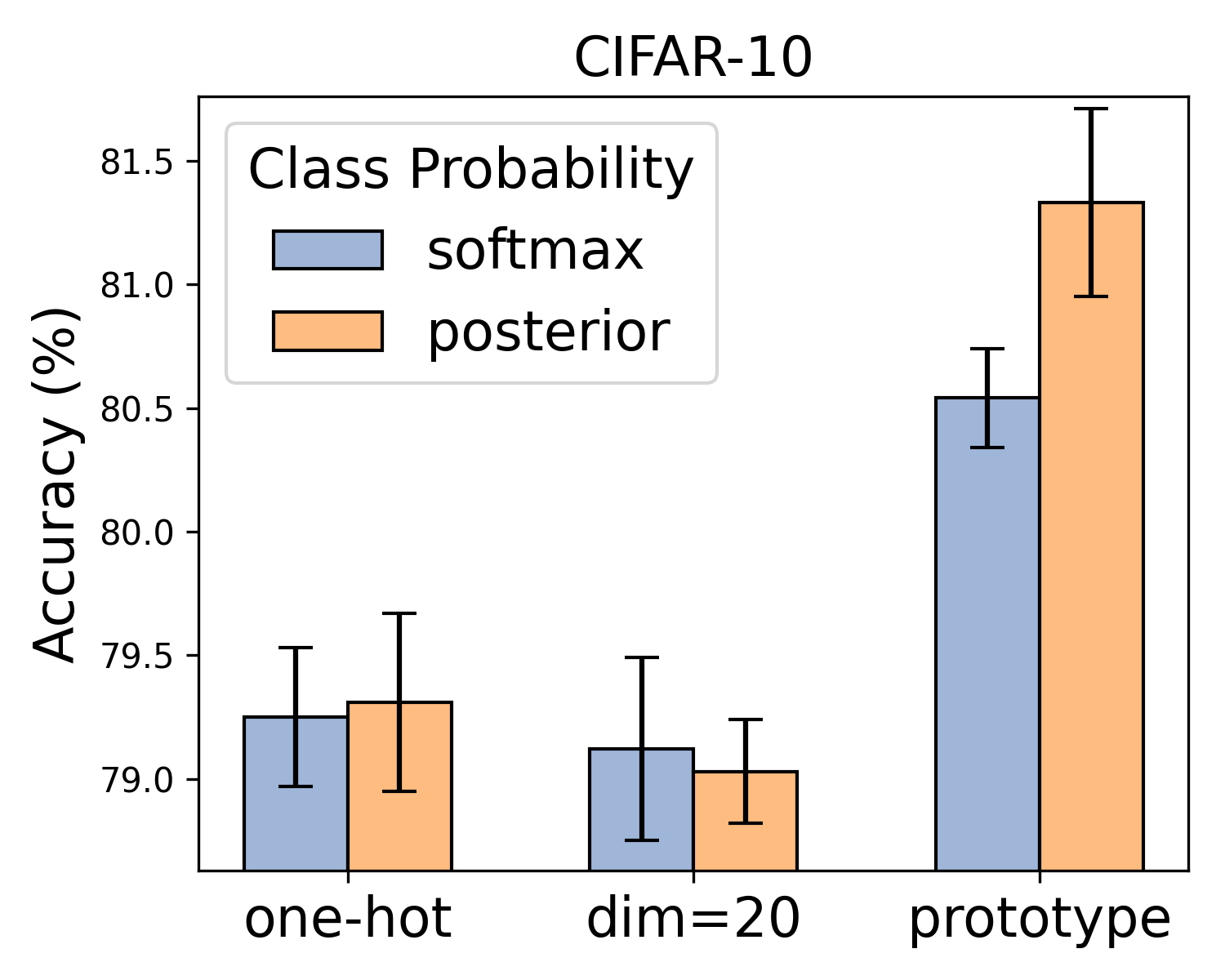} 
        \end{minipage}%
        \begin{minipage}{0.31\textwidth} 
            \centering
            \includegraphics[width=\linewidth]{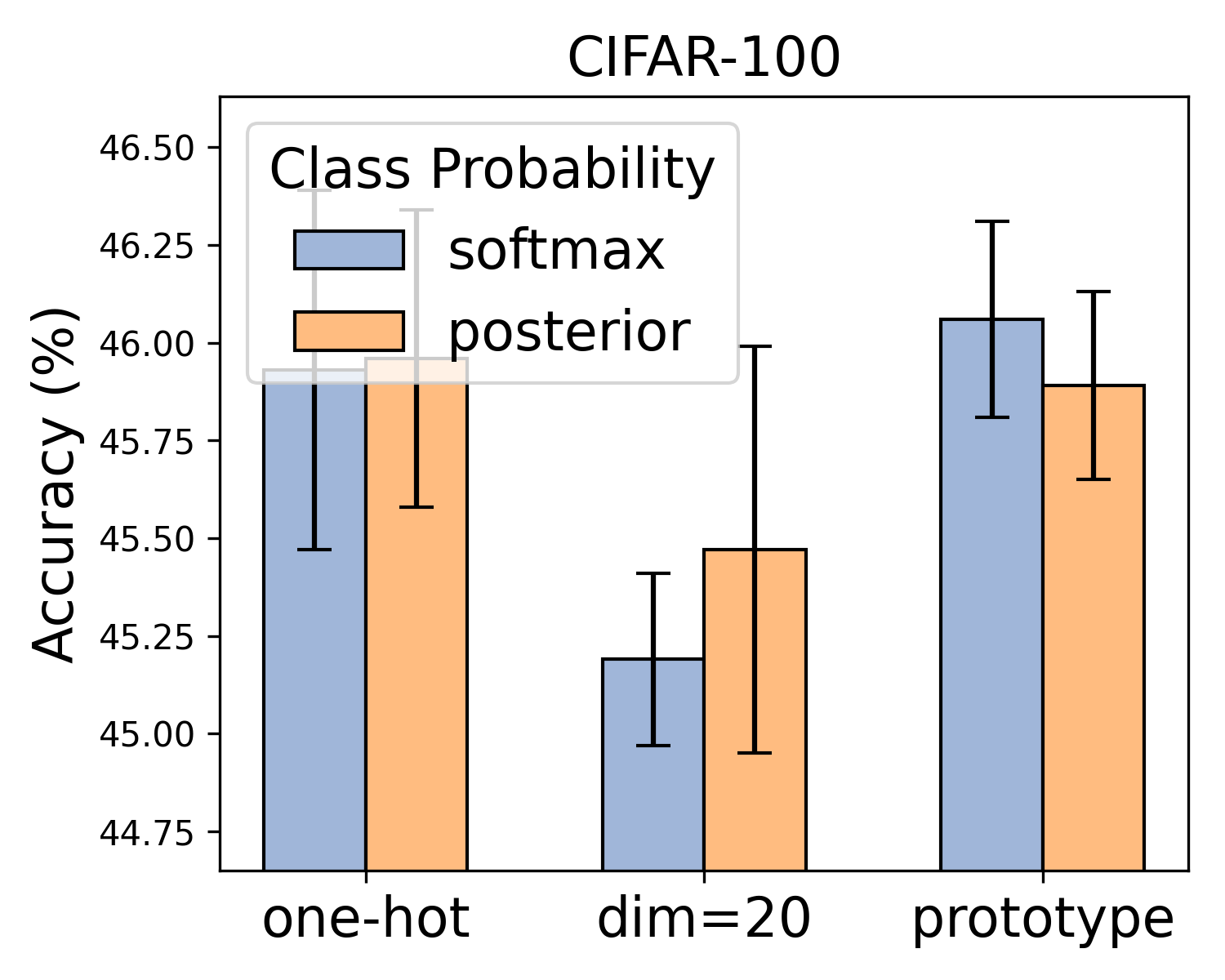} 
        \end{minipage}
        \caption{Test accuracy (\%) comparison for two parameterisations of class probabilities $\hat{p}_{\theta_{\textrm{out}}}(y|z_T)$ in Equation~\ref{eq:noprop_dt}, using either softmax or posterior probability.} 
        \label{fig:class_probabilities}
    \end{figure}

    \begin{figure}[htbp]
        \centering
        \begin{minipage}{0.32\textwidth} 
            \centering
            \includegraphics[width=\linewidth]{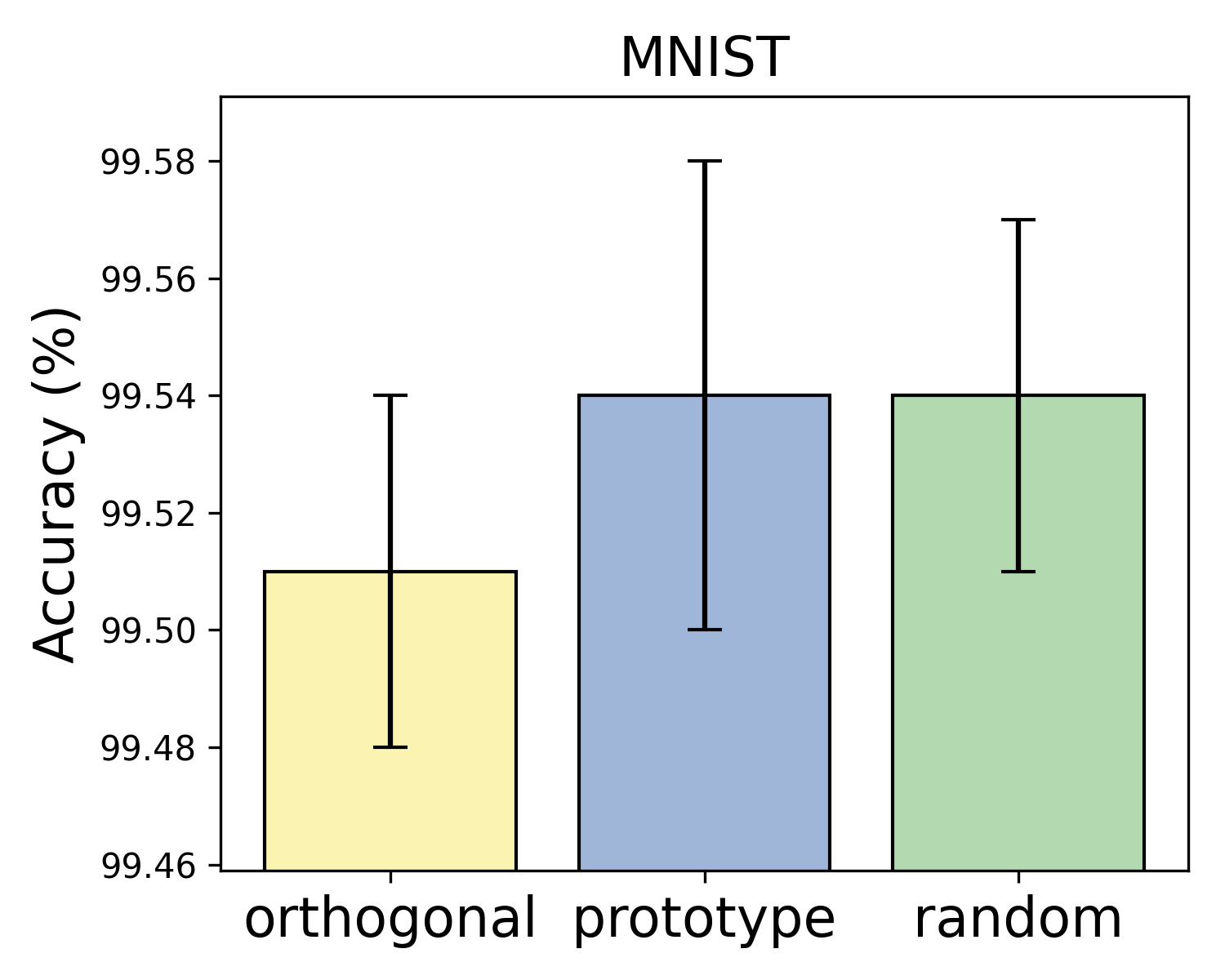} 
        \end{minipage}%
        \begin{minipage}{0.32\textwidth} 
            \centering
            \includegraphics[width=\linewidth]{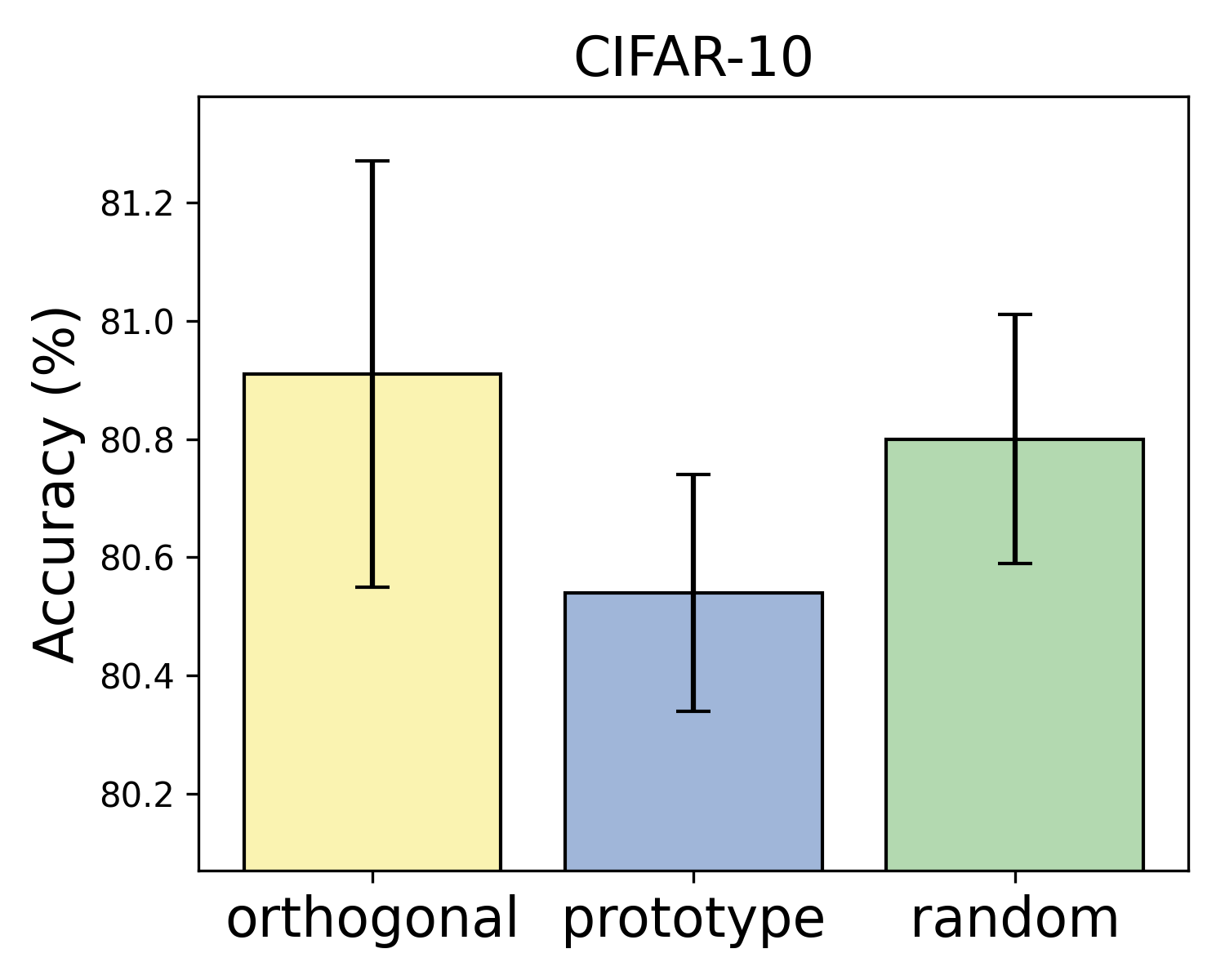} 
        \end{minipage}%
        \begin{minipage}{0.32\textwidth} 
            \centering
            \includegraphics[width=\linewidth]{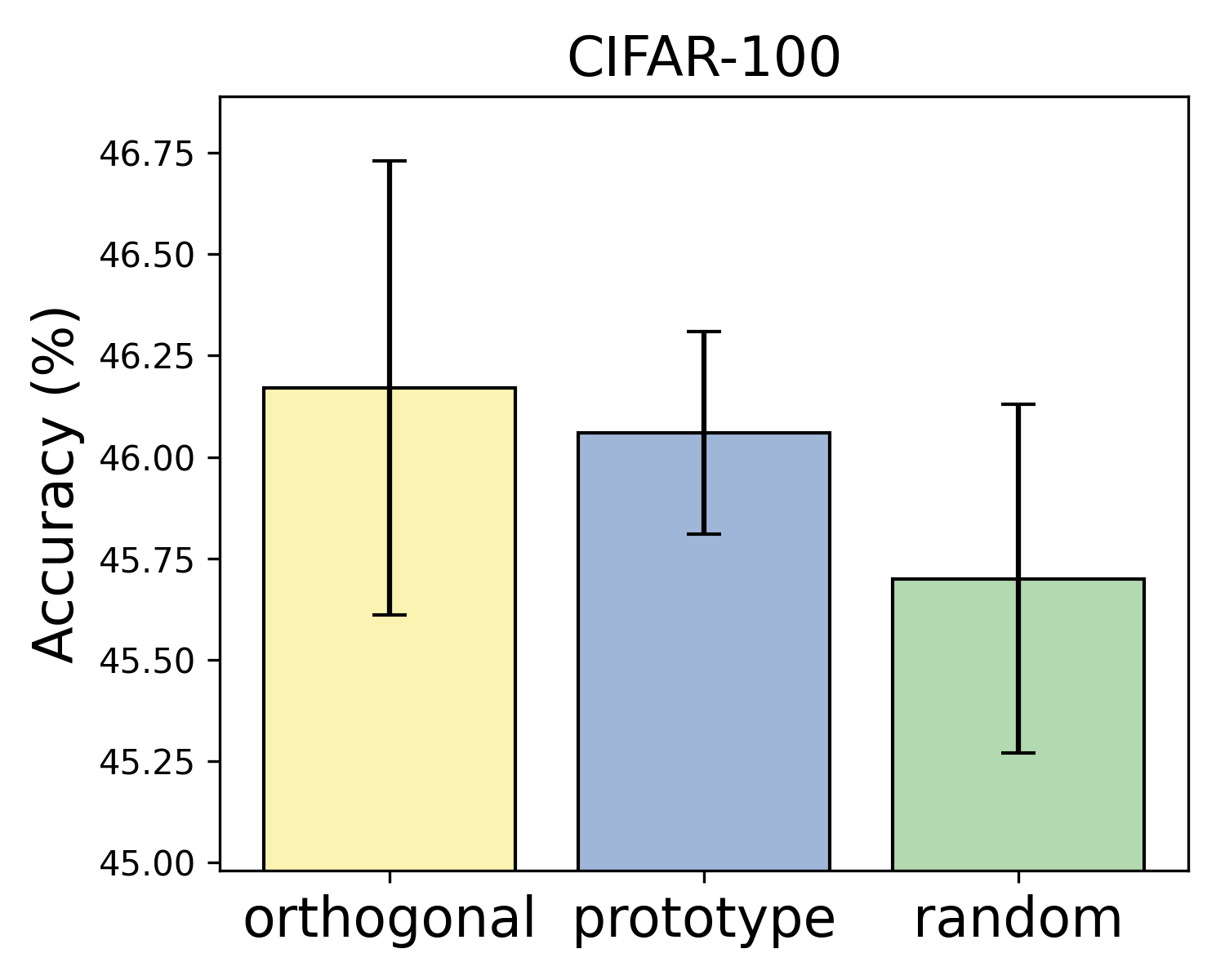} 
        \end{minipage}
        \caption{Test accuracy (\%) comparison for different initializations of the class embedding matrix $W_{\text{Embed}}$ when the class embedding dimension matches the image dimension. The initializations considered are random matrices, orthogonal matrices, and prototype images.
        } 
        \label{fig:initializations}
    \end{figure}

\begin{table}[]
    \centering
    \begin{tabular}{@{}lllll@{}}
    \toprule
    Dataset                        & One-hot & Prototype & Label dim & Params \\ \midrule
    \multicolumn{5}{c}{Discrete-time}                                         \\ \midrule
    \multirow[t]{3}{*}{MNIST}         & Yes     & No        & 10        & 0.92M  \\
                                   & No      & No        & 20        & 0.92M  \\
                                   & No      & Yes       & 784       & 1.40M  \\
    \multirow[t]{3}{*}{CIFAR-10}      & Yes     & No        & 10        & 1.22M  \\
                                   & No      & No        & 20        & 1.22M  \\
                                   & No      & Yes       & 3072      & 1.99M  \\
    \multirow[t]{3}{*}{CIFAR-100}     & Yes     & No        & 100       & 1.25M  \\
                                   & No      & No        & 20        & 1.23M  \\
                                   & No      & Yes       & 3072      & 2.00M  \\
    \multirow[t]{2}{*}{Tiny ImageNet} & Yes     & No        & 200       & 2.45M  \\
                                   & No      & Yes       & 12288     & 4.30M  \\ \midrule
    \multicolumn{5}{c}{Continuous-time}                                       \\ \midrule
    \multirow[t]{3}{*}{MNIST}         & Yes     & No        & 10        & 0.87M  \\
                                   & No      & No        & 20        & 0.87M  \\
                                   & No      & Yes       & 784       & 1.48M  \\
    \multirow[t]{3}{*}{CIFAR-10}      & Yes     & No        & 10        & 1.16M  \\
                                   & No      & No        & 20        & 1.17M  \\
                                   & No      & Yes       & 3072      & 2.07M  \\
    \multirow[t]{3}{*}{CIFAR-100}     & Yes     & No        & 100       & 1.20M  \\
                                   & No      & No        & 20        & 1.18M  \\
                                   & No      & Yes       & 3072      & 2.08M  \\ \bottomrule
    \end{tabular}
    \caption{Model parameters.}
    \end{table}

\end{document}